\documentclass[journal]{IEEEtran}
\usepackage[percent]{overpic}

\usepackage{tabu}

\usepackage{multirow}
\def\mr{\multirow}
\def\mc{\multicolumn}

\usepackage[overload]{empheq}             

\usepackage[table,xcdraw]{xcolor}
\definecolor{dg}{rgb}{0.1, 0.6, 0.2}       
\definecolor{b}{rgb}{0.0, 0.0, 1}          
\usepackage{colortbl}                      

\usepackage{epsfig}
\usepackage{float}
\usepackage{color}
\usepackage{url}

\usepackage{algorithm}      
\usepackage{algpseudocode}

\usepackage{amssymb, amsfonts}
\usepackage{amsmath}
\usepackage{mathrsfs}

\usepackage{amsthm}
\usepackage{mathtools}

\let\labelindent\relax
\usepackage{enumitem}       
\setenumerate[enumerate]{align=left}

\usepackage{graphicx}
\usepackage{subcaption}
\usepackage{caption}

\usepackage{cite}

\makeatletter
\let\NAT@parse\undefined
\makeatother
\usepackage{hyperref} 

\usepackage{balance}

\usepackage{bm}

\usepackage{rotating}

\newcommand{\floor}[1]{\left\lfloor #1 \right\rfloor}

\newcommand{\norm}[1]{\left\lVert#1\right\rVert}

\newcommand{\Mint}{%
  \mathrlap{\mathop{\phantom{\int}}\limits_{\!\mathcal{M}}}%
  \!\int
}

\makeatletter
\newlength\tmp@\newlength\t@mp
\newcommand{\comp}[3]
  {\mathop{ \settowidth\tmp@{$\displaystyle\mathop{#1}^{#3}_{#2}$}
  \hbox to \tmp@{\hss \settowidth\t@mp{$\displaystyle #1$}\setlength\t@mp{.45\t@mp}
  $\displaystyle\mathop{#1}^{\hspace\t@mp #3}_{\hspace{-\t@mp}#2}$
  \hss} }}
\makeatother

\newcommand{\Int}[2]
{\int_{#1}^{#2}}

\hypersetup{
    colorlinks=true,
    linkcolor=blue,
    filecolor=blue,      
    urlcolor=blue,
    citecolor=blue,
}

\def\a{\alpha}
\def\b{\beta}
\def\d{\delta}

\def\g{\gamma}

\def\o{\omega}
\def\s{\sigma}

\def\D{\Delta}

\def\R{\mathbb{R}}
\def\N{\mathbb{N}}

\def\l{\left}
\def\r{\right}

\def\quat{\mathbf{q}}

\def\pos{\mathbf{p}}
\def\vel{\mathbf{v}}
\def\accel{\mathbf{a}}

\newcommand{\sksym}[1]{\floor{#1}_{\times}}
\newcommand{\fr}[1]{\texttt{#1}}
\newcommand{\qmatL}[1]{\floor{#1}_\ell}
\newcommand{\qmatR}[1]{\floor{#1}_r}
\newcommand{\vbf}[1]{\bm{\mathbf{#1}}}
\def\brc{\mathbf{brc}}
\def\calE{\mathcal{E}}

\def\resi{\bm{r}}
\def\Lift{\mathcal{L}}
\def\Jcb{\vbf{J}}

\def\bias{\mathbf{b}}
\def\noise{\bm{\eta}}
\def\rot{\mathbf{R}}
\def\tf{\mathbf{T}}
\def\trans{\mathbf{t}}
\def\SO{\mathrm{SO(3)}}
\def\so{\mathfrak{so}(3)}
\def\SE{\mathrm{SE(3)}}
\def\Rtoq{\mathcal{Q}}
\def\qtoR{\mathcal{R}}
\def\Exp{\mathrm{Exp}}
\def\Log{\mathrm{Log}}
\def\Hrj{\mathcal{H}}
\def\Omg{{\bm\Omega}}
\def\cov{\mathrm{Cov}}

\def\ident{\mathbf{I}}
\def\zero{\mathbf{0}}

\def\Dt{ {\D t} }
\def\dt{ {\d t} }
\def\wrt{\text{w.r.t. }}

\def\pia{{\vbf{\a}}}
\def\pib{{\vbf{\b}}}
\def\pig{{\vbf{\g}}}
\def\pith{{\vbf{\theta}}}

\def\X{\mathcal{X}}

\def\U{\mathcal{U}}
\def\I{\mathcal{I}}
\def\V{\mathcal{V}}
\def\M{\mathcal{M}}

\def\angvel{\bm{\omega}}
\def\accel{\mathbf{a}}
\def\grav{\mathbf{g}}
\def\eulvec{\bm{\phi}}

\def\fA{\fr{A}}
\def\fB{\fr{B}}
\def\fC{\fr{C}}
\def\fL{\fr{L}}
\def\fW{\fr{W}}
\def\fI{\fr{I}}

\newtheorem{rem}{Remark}

\begin{document}

\title{VIRAL-Fusion: A Visual-Inertial-Ranging-Lidar Sensor Fusion Approach}

\author{
      Thien-Minh Nguyen,
      Shenghai Yuan,
      Muqing Cao,
      Yang Lyu,\\
      Thien Hoang Nguyen,
      and Lihua Xie, \IEEEmembership{Fellow,~IEEE}
\thanks{
The work is supported by National Research Foundation (NRF) Singapore, ST Engineering-NTU Corporate Lab under its NRF Corporate Lab@ University Scheme.}
\thanks{The authors are with School of Electrical and Electronic Engineering, Nanyang Technological University, 50 Nanyang Avenue, Singapore 639798.}
\thanks{ Email for correspondence: thienminh.npn@ieee.org (Thien-Minh Nguyen)}
}


\maketitle

\begin{abstract}
In recent years, Onboard Self Localization (OSL) methods based on cameras or Lidar have achieved many significant progresses. However, some issues such as estimation drift and feature-dependence still remain inherent limitations. On the other hand, infrastructure-based methods can generally overcome these issues, but at the expense of some installation cost. This poses an interesting problem of how to effectively combine these methods, so as to achieve localization with long-term \textit{consistency} as well as \textit{flexibility} compared to any single method. To this end, we propose a comprehensive optimization-based estimator for 15-dimensional state of an Unmanned Aerial Vehicle (UAV), fusing data from an extensive set of sensors: inertial measurement units (IMUs), Ultra-Wideband (UWB) ranging sensors, and multiple onboard Visual-Inertial and Lidar odometry subsystems. In essence, a sliding window is used to formulate a sequence of robot poses, where relative rotational and translational constraints between these poses are observed in the IMU preintegration and OSL observations, while orientation and position are coupled in \textit{body-offset} UWB range observations. An optimization-based approach is developed to estimate the trajectory of the robot in this sliding window.
We evaluate the performance of the proposed scheme in multiple scenarios, including experiments on public datasets, high-fidelity graphical-physical simulator, and field-collected data from UAV flight tests. {\color{black} The result demonstrates that our integrated localization method can effectively resolve the drift issue, while incurring minimal installation requirements.}
\end{abstract}


\IEEEpeerreviewmaketitle

\section{Introduction} \label{sec: intro}
\IEEEPARstart{S}{tate} estimation is a fundamental task in developing autonomous mobile robots. The problem is even more critical when dealing with unmanned aerial vehicles (UAVs), since they can be quite easily destroyed when system failures occur, yet limited payload capacity also often prevents one from including so many backup systems on the platform.
Thus, depending on the specific scenario, one has to carefully assess the priorities on reliability, accuracy, flexibility, and then settle on appropriate set of sensors and fusion techniques for the targeted application.

With the goal of achieving a reliable localization system for UAV-based inspection applications of 3D structures, in this work, we study a general sensor fusion scheme for fusing measurements from Inertial Measurement Units (IMUs), camera/Lidar-based Onboard Self-Localization (OSL) systems, and UWB, hence the name \textit{\underline{V}isual-\underline{I}nertial-\underline{Ra}nging-\underline{L}idar Sensor Fusion} framework.
Indeed, many developers would find OSL the first choice for the localization task. Thanks to many advances in hardware and software design, it has become quite popular and ubiquitous in research and industry in recent years. Currently, one can find quite reliable and compact off-the-shelf camera/Lidar sensors on the market, along with many open-sourced software software packages to be used with these sensors \cite{shen20113d, bloesch2015robust, engel2014lsd, weinstein2018visual, forster2014svo, qin2017vins, mur2017orb, zhang2018laser, ye2019tightly, shan2020liosam, suleiman2019navion}.
However, besides the sensitivity to lighting condition and availability of features in the environment, OSL systems have two characteristics that one needs to pay attention to. First, OSL data are often referenced to a local coordinate frame that coincides with the initial pose, and second, there will be significant estimation drift over a long period of time. These issues are quite adversely challenging in inspection or surveillance applications, when predefined trajectories need to be executed in reference to a chosen so-called world frame. Thus, some small misalignment of the initial pose can eventually lead to large errors towards the end of the operation.

Therefore, in this work we investigate the fusion of OSL data with range measurements to some UWB anchors, which can be deployed in a field to define such a global frame. To keep the system easy to use, we also employ very simple deployment strategies.
Moreover, we also innovate the use of UWB with a \textit{body-offset ranging scheme}, where orientation is also coupled with the range measurements besides the positions. Such coupling ensures that both position and orientation estimates do not drift during the operation.
Besides these advantages, since it is well known that most VIOs can potentially "lose track", relative observations that couple consecutive poses are also constructed from IMU data and fused with UWB measurements to ensure the continuity of estimation. Note that despite its high reliability and inexpensiveness, the integration of the IMU is not a trivial task, as one has to carefully derive the correct kinematic model, handle the biases, and on top of that properly synchronize the high-rate IMU data with other sensors. These issues are successfully addressed in this paper using a synchronization scheme inspired by the image-IMU synchronization process in state of the art VIO \cite{qin2017vins, qin2018online} and Lidar-inertial systems \cite{zhang2018laser, ye2019tightly, shan2020liosam}. Hence, IMU, OSL and UWB measurements are effectively fused together to provide accurate and reliable localization for the UAV application.

\begin{table*}[!h]
    \setlength{\tabcolsep}{3.75pt}
    \centering
    \renewcommand{\arraystretch}{1.1}
	\caption{Related works and their classifications.}
	\label{tab: related works}
    \begin{tabu} to \textwidth {c||c|c|c|c||c|c|c}
    \hline\hline
                                &\mc{4}{c||}{Sensor types}                              &\mc{2}{c|}{UWB Utilization}&\mr{2}{*}{Fusion Method}\\\cline{2-7}
    \mr{-2}{*}{Related works}   &UWB           &IMU            &Camera      &Lidar      &State Coupling       &Localization Type   &\\\hline
    \multicolumn{1}{l||}
    {UWB only: \cite{guo2016ultra, nguyen2016ultra, tiemann2017scalable}}
                                &\checkmark    &               &-           &-          &Position      &Absolute    &EKF\\ \hline
    \multicolumn{1}{l||}
    {UWB-IMU: \cite{mueller2015fusing, li2018accurate}}
                                &\checkmark    &\checkmark     &-           &-          &Position      &Absolute    &EKF\\ \hline
    \mc{1}{l||}{Range-only Graph Optimization: \cite{fang2018model, fang2019graph}}
                                &\checkmark    &-              &-           &-          &Position     &Absolute     &Optimization\\ \hline
    
    \mc{1}{l||}{UWB-aided Visual SLAM (VSLAM) \cite{wang2017ultra}}
                                &\checkmark    &-              &\checkmark  &-          &Position     &Absolute    &Optimization\\ \hline
    
    \mc{1}{l||}{UWB-aided Monocular VSLAM: \cite{nguyen2019loosely, nguyen2019tightly, nguyen2020tightly}}
                                &\checkmark    &-              &\checkmark  &-          &Position      &Relative    &Optimization\\ \hline
    
    \mc{1}{l||}{VIR-SLAM: \cite{cao2020vir}}
                                &\checkmark    &\checkmark     &\checkmark  &-          &Position      &Relative    &Optimization\\ \hline
    
    \mc{1}{l||}{UWB/Lidar 2D SLAM: \cite{song2019uwb}}
                                &\checkmark    &-              &-           &\checkmark &2D Position   &Relative    &Optimization\\ \hline
    
    \mc{1}{l||}{Ours}           &\checkmark    &\checkmark     &\checkmark  &\checkmark &Position, Orientation, Velocity
                                                                                                       &Absolute    &Optimization\\ \hline
                                
    \end{tabu}
\end{table*}

The remaining of this paper is organized as follows. In Section \ref{sec: related works}, we review some of the most related previous works and highlight our contributions. Section \ref{sec: preliminary} provides several basic concepts that will be employed in the later parts. Section \ref{sec: problem description} provides an overview of the approach to the estimation problem. Section \ref{sec: cost factors} goes into detail on how to construct the cost function.
Section \ref{sec: experiment} presents the experiment results. Finally, Section \ref{sec: conclusion} concludes our work.

\section{Related Works} \label{sec: related works}

To the best of our knowledge, this work presents an original and comprehensive sensor fusion scheme in terms of the set of sensors, their utilization for estimation, and the fusion approach. Tab. \ref{tab: related works} summarizes and compares the most related works and ours with respect to these aspects. Note that we only consider the works that employ UWBs in combination with other types of sensors.

As Tab. \ref{tab: related works} presents, over the last five years, many researchers have investigated the use of UWB for UAV localization as a standalone method \cite{mueller2015fusing, guo2016ultra, nguyen2016ultra, tiemann2017scalable, paredes20183d}, or in combination with IMU \cite{mueller2015fusing, li2018accurate}. These works have confirmed the effectiveness of UWB for localization in the mid-range scale (from 10 m to 100 m) \cite{mautz2012indoor}. 
Since these early works use Extended Kalman Filter (EKF), the fusion of UWB with relative motion information from OSL system can be quite cumbersome. In an early attempt to find an alternative to EKF, in \cite{fang2018model, fang2019graph} the authors proposed optimizing a cost function comprising of ranging and smoothness factors on a sliding window. This optimization process provides a sequence of pose estimates that best fit the available measurements. Note that the smoothness factors in this formulation are some artificially designed relative factors, i.e. factors that couple two consecutive states in the sliding window, to keep the problem well-posed \cite{fang2019graph}. As the term smoothness suggests, these factors penalize the solutions where consecutive positions are too close or too far from each other, and assume some conditions on the constant velocity and maximum speed of the robot. Hence the method may only be effective when the robot moves at a low speed. In \cite{wang2017ultra}, the authors introduced new relative factors from a Visual SLAM system to the cost function, together with the range and smoothness constraints, and demonstrated the improved localization accuracy compared with purely visual methods.

In the aforementioned methods, it is assumed that UWB anchor positions are known and localization is with respect to the coordinate system specified by the anchor setup. These often require at least four anchors to ensure a unique solution. On the other hand, in \cite{nguyen2019loosely, nguyen2019tightly, nguyen2020tightly, cao2020vir, song2019uwb}, ranging to a single anchor or several unknown anchors are used to aid the OSL systems. More specifically, the works in \cite{nguyen2019loosely, nguyen2019tightly, nguyen2020tightly} employ UWB ranges to estimate the scale factors, which are ambiguous in monocular Visual SLAM systems. Whereas in \cite{cao2020vir}, ranging to a single anchor and among a group of robots are used to reduce the drift of the OSL data. In \cite{song2019uwb}, the authors proposed a 2D UWB/Lidar SLAM problem, where anchor-robot and anchor-anchor ranges along with Lidar scans are used to simultaneously estimate the anchor and robot positions. It should be noted that the position estimates from these methods are still in an arbitrary coordinate frame, and therefore not conducive for executing trajectories relative to some fixed frame of references, which is typical in inspection tasks.

Compared with the previous works, the first contribution of our work is the fusion of a comprehensive set of sensors, namely UWB, IMU, VIO and LOAM (in fact, multiple cameras and Lidars are employed instead of one for each type in our system). Indeed, the integration of such a comprehensive set of sensors prompts a need to design a more comprehensive sensor fusion scheme, whereas previous schemes cannot be easily applied. We will discuss this issue in more details in Section \ref{sec: problem description}.
Obviously, the first benefit of using more sensors is to improve the reliability of the localization, especially when OSL methods are involved, as they can lose track and destabilise the localization process. Hence, by employing multiple OSL methods together, we can more reliably detect anomaly in one method and temporarily ignore its data. Moreover, our work also employs a tightly-coupled fusion of IMU-preintegration and UWB ranges in the optimization approach, hence even when all OSL information is lost, localization can still be maintained. This approach is also better than the previous works' use of smoothness factors, as no strict assumption on maximum velocity is needed.

Moreover, we notice that the previous works only features the coupling of UWB measurements with position state. As mentioned, our method introduces a body-offset ranging scheme plus an observation model that couple both position, orientation and velocity with the range measurements. Thus, not only the use of range factors can reduce the drift in position, but also orientation and velocity. These improvements will be discussed in details in Section \ref{sec: experiment}.

Finally, it should be noted that the derivation of Jacobians that is needed for optimization in previous works were often glossed over, or approximated with numerical methods \cite{kummerle2011g}. Hence their methods would have the potential of facing numerical stability. In contrast, since the analytical forms of the Jacobians can ensure more accurate and fast computations, in this paper we also provide a detailed mathematical derivations of the Jacobians for the IMU, OSL and body-offset UWB ranges, backed up by a comprehensive and succinct self-contained theoretical framework. Our system is successfully implemented on the ceres optimization framework \cite{ceres-solver}, which are also used by a majority of current state of the art VIO and LOAM systems \cite{qin2018vins, qin2018online, zhang2018laser, ye2019tightly, campos2020orb}.

\section{Preliminaries} \label{sec: preliminary}

\subsection{Notations}
In this article we respectively use $\R$, $ \N $, and $ \N^+$ to denote the sets of natural numbers, non-negative integers, and positive integers. 
$ \R^m $ denotes the $ m $-dimension real vector space with $ m \in \N $.

We use $(\cdot)^\top$ to denote the transpose of a vector or matrix under $(\cdot)$.
For a vector $ \vbf{x} \in \R^m $, $\norm{\vbf{x}}$ stands for its Euclidean norm, and $\norm{\vbf{x}}_{\vbf{G}}^2$ is  short-hand for $\norm{\vbf{x}}_{\vbf{G}}^2 = \vbf{x}^\top \vbf{G} \vbf{x}$. For a matrix $\vbf{A}$, $\brc_m(\vbf{A})$ returns the $m\times m$ block at bottom right corner of $\vbf{A}$, e.g. $\brc_1\l(\begin{bmatrix}5 &7\\ 4 &3\end{bmatrix}\r) = \l[3\r]$.

To make it explicit that a vector $\mathbf{v}$ is \wrt a coordinate frame $\{\fr{A}\}$, we attach a left superscript ${}^{\fr{A}}$ to $\vel$, e.g. ${}^\fr{A}\mathbf{v}$. A rotation matrix and transform matrix between two reference frames are denoted with the frames attached as the left-hand-side superscript and subscript, e.g. ${}^\fA_\fB\rot$ and ${^\fA_\fB\tf}$ are the rotation and transform matrices from frame $\{\fA\}$ and $\{\fB\}$, respectively. Morover, left-hand-side subscript ${}_{\fB}(\cdot)$ (``$\fB$" is for body frame) is implied when none is specified, e.g., ${}^{\fL}\quat_k$ implies the quaternion representing the rotation from the local frame $\fL$ to the body frame ${\fB}$ at time $t_k$, i.e, ${}^{\fL}_{\fB_k}\quat$.

For a list of vectors $\vel_1, \vel_2 \dots, \vel_n$ of various dimensions (including scalar), we use the tuple notation $(\vel_1, \vel_2 \dots, \vel_n)$ as a short-hand for $[\vel_1^\top, \vel_2^\top \dots, \vel_n^\top]^\top$.

Finally, throughout the paper, super/subscripts may be occasionally omitted for conciseness, or inserted for clarity, when deemed unambiguous in its context.

\subsection{Geometric concepts}

\subsubsection{Special Orthogonal Group} $\SO$ describes the group of 3D rotations, which are formally defined as
\begin{equation*}
    \SO = \{\rot \in \R^{3\times3} \colon \rot^\top\rot = \ident,\ \det(\rot) = 1 \}.
\end{equation*}
On this algebraic group, the \textit{binary operation} \cite{chirikjian2011stochastic} representing the composition of rotations is defined as $\rot_1 \circ \rot_2 = \rot_1\rot_2$. By definition it can also be seen that the inverse of a rotation matrix is defined as $\rot^{-1} = \rot^\top$.

In a practical sense, a rotation matrix ${}^\fA_\fB\rot$ helps convert the coordinate of a vector in the frame $\{\fB\}$ to that in the frame $\{\fA\}$, e.g. a vector $\vbf{v}$ whose coordinate in $\{\fB\}$ is ${}^\fB\vbf{v}$ has its coordinates in $\{\fA\}$ as ${}^\fA\vel = {}^\fA_\fB\rot {}^\fB\vbf{v}$.

\subsubsection{Skew-symmetric matrix}
We define the skew-symetric matrix of a vector $\mathbf{a} = [a_x,\ a_y,\ a_z]^\top \in \R^3$ as
\begin{equation}
    \sksym{\mathbf{a}} = \l[\begin{array}{ccc} 0   &-a_z &a_y\\
                                               a_z &0    &-a_x\\
                                              -a_y &a_x &0\end{array}\r],
\end{equation}
which helps define the cross product $\times$ between two vectors:
\begin{equation} \label{eq: skew and cross}
    \vbf{a}\times\vbf{b} = \sksym{\vbf{a}}\vbf{b} = - \sksym{\vbf{b}}\vbf{a}.
\end{equation}

The space of skew symetric matrices is a \textit{Lie algebra} associated with $\SO$, denoted as $\so$.
We also define the following mappings between $\R^3$ and $\so$ that will be used in later formulations:
\begin{align}
    (\cdot)^\wedge &\colon \R^3 \to \so;\quad \vbf{\tau}^\wedge = \sksym{\vbf{\tau}};\\
    (\cdot)^\lor &\colon  \so \to \R^3 ;\quad \sksym{\vbf{\tau}}^\lor = \vbf{\tau}.
\end{align}

The Lie algebra is called the "infinitestimal generator" of $\SO$ as one can map an element of $\so$ into $\SO$ via the matrix exponential:
\begin{equation*}
    \forall \vbf{\tau}^\wedge \in \so
    \colon
    \exp(\vbf{\tau}^\wedge) \triangleq \ident + \sum_{k=1}^\infty\frac{1}{k!}\sksym{\vbf{\tau}}^k = \rot \in \SO.
\end{equation*}
Following \cite{sola2018micro}, a closed-form solution of the above matrix exponential is provided as follows
\begin{align}\label{eq: rodrigues formula}
    &\Exp(\eulvec) = \ident + \frac{\sin(\norm{\eulvec})}{\norm{\eulvec}}\sksym{\eulvec} + \frac{1 - \cos(\norm{\eulvec})}{\norm{\eulvec}^2}\sksym{\eulvec}^2,
\end{align}
where $\Exp(\eulvec) \triangleq \exp(\eulvec^\wedge)$.
This equation is commonly known as \textit{Rodrigues' formula}, and $\eulvec$ is the axis–angle representation of a rotation matrix $\rot = \Exp(\eulvec)$, also called \textit{rotation vector}. The inverse mapping from a rotation matrix to a rotation vector is defined as
\begin{align}
    &\Log(\rot) = \eulvec = \frac{ \norm{\eulvec}}{2\sin(\norm{\eulvec})}(\rot - \rot^\top)^\lor,\nonumber\\
    &\norm{\eulvec} = \cos^{-1}\l(\frac{\mathrm{Tr}(\rot)-1}{2}\r), \label{eq: rodrigues formula inverse}
\end{align}
where $\mathrm{Tr}(\rot)$ denotes the trace of the matrix $\rot$.

From the definitions of rotation matrix and skew symmetric matrix, we can verify the following identities for all $\rot \in \SE$ and $\eulvec \in \R^3$: 
\begin{align}
    & \sksym{\rot\eulvec} = \rot\sksym{\eulvec}\rot^\top,\ \Exp(\eulvec)\rot = \rot\Exp(\rot^\top\eulvec). \label{eq: ER=RE}
\end{align}

We also have the first order approximation for the mapping \eqref{eq: rodrigues formula} under a small disturbance $\d\phi$:
\begin{align}
    &\Exp(\eulvec + \d\vbf{\phi}) \simeq \Exp(\eulvec)\Exp\l(\Hrj(\eulvec)\d\vbf{\phi}\r), \label{eq: first order approx of rodrigues formula}
    \\
    &\Exp(\eulvec)\Exp(\d\vbf{\phi}) \simeq \Exp\l(\eulvec + \Hrj^{-1}(\eulvec)\d\vbf{\phi}\r),
    \label{eq: first order approx of rodrigues formula inverse}
\end{align}
where $\Hrj(\eulvec)$ is termed the \textit{right Jacobian} of $\SO$ \cite{chirikjian2011stochastic}, and is formally defined as:
\begin{equation}\label{eq: right jacobian definition}
     \Hrj(\eulvec) = \dfrac{\partial\Log\l[\Exp(\eulvec)^{-1}\Exp(\eulvec + \pith)\r]}{\partial \pith}\Bigg|_{\pith = 0}
\end{equation}
The closed-form of $\Hrj(\eulvec)$ and its inverse are given as follows
\begin{align*}
    &\Hrj(\eulvec) = \ident - \frac{1 - \cos(\norm{\eulvec})}{\norm{\eulvec}^2}\sksym{\eulvec} + \frac{\norm{\eulvec} - \sin(\norm{\eulvec})}{\norm{\eulvec}^3}\sksym{\eulvec}^2.\\
    &\Hrj^{-1}(\eulvec) = \ident + \frac{1}{2}\sksym{\eulvec} +\l(\frac{1}{\norm{\eulvec}^2} - \frac{1 + \cos(\norm{\eulvec})}{2\norm{\eulvec}\sin(\norm{\eulvec})}\r)\sksym{\eulvec}^2.
\end{align*}

Note that when $\eulvec$ is  small, then $\Hrj(\phi) \simeq \ident$. Thus for two small disturbances $\d\vbf{\phi}_1$, $\d\vbf{\phi}_2$ we can have:
\begin{align}
    \Exp(\d\vbf{\phi}_1 + \d\vbf{\phi}_2) \simeq \Exp(\d\vbf{\phi}_1)\Exp\l(\d\vbf{\phi}_2\r).
    \label{eq: first order approx two disturbance}
\end{align}
This result will later be used to derive the equations for propagating the covariance of the IMU preintegration measurements.
\subsubsection{Quaternion}
A quaternion $\quat$ is a four-dimensional complex number comprising of a scalar part $q_w \in \R$, and a vector part $\quat_v \in \R^3$:
\begin{align}
    &\quat = q_w + \vbf{i}q_x + \vbf{j}q_y + \vbf{k}q_z = (q_w,\ \quat_v),\\
    &q_w,\  q_x,\  q_y,\  q_z \in \R,\ \quat_v = (q_x,\quad q_y,\quad q_z) \in \R^3.
\end{align}
A \textit{unit quaternions}, i.e. quaternion $\quat$ where $\norm{\quat} = 1$, can be used to represent a rotation. The formulae to convert a unit quaternion $\quat$ to a rotation matrix $\rot = \{a_{ij}\},\ i, j \in \{1, 2, 3\}$ and vice versa are recalled from \cite{huang2020online} as follows:
\begin{align}
    &\qtoR(\quat) = (2q_w^2 -1)\ident_{3\times3} + 2q_w\sksym{\quat_v}+2\quat_v\quat_v^\top, \label{eq: quat to rot}\\
    &\Rtoq(\rot)  = (q_w, q_x, q_y, q_z), q_w = \frac{\sqrt{a_{11} + a_{22} + a_{33} + 1}}{2},\\
    &q_x = \dfrac{a_{32} - a_{23}}{4q_w},\ q_y = \dfrac{a_{13} - a_{31}}{4q_w},\ q_z = \dfrac{a_{21} - a_{12}}{4q_w}. \label{eq: rot to quat} 
\end{align}

We also have the following formula for the compound mapping $\calE(\eulvec) = \Rtoq(\Exp(\eulvec))$:
\begin{equation} \label{eq: vec to quat}
    \forall \eulvec \in \R^3 \colon
    \calE(\eulvec) = 
            \l(\cos\l(\dfrac{\norm{\eulvec}}{2}\r),\ 
            \dfrac{\eulvec}{\norm{\eulvec}}\sin\l(\dfrac{\norm{\eulvec}}{2}\r)\r)
\end{equation}

Given two rotation matrices ${}^\fA_\fB\rot$, ${}^\fB_\fC\rot$ and their corresponding quaternions ${}^\fA_\fB\quat$, ${}^\fA_\fB\quat$, the rotation and quaternion composition operations can be defined via the matrix multiplications, i.e.:
\begin{gather}
    {}^\fA_\fC\rot = {}^\fA_\fB\rot {}^\fB_\fC\rot \Leftrightarrow \quat \circ \mathbf{p} = \qmatL{\quat}\mathbf{p} = \qmatR{\mathbf{p}}\quat,\\
    \qmatL{\quat} = \begin{bmatrix} q_w        &-\vbf{q}_v^\top\\
                                         \vbf{q}_v  &q_w \vbf{I} + \sksym{\vbf{q}_v} \end{bmatrix},\\
    \qmatR{\quat} = \begin{bmatrix} q_w        &-\vbf{q}_v^\top\\
                                         \vbf{q}_v  &q_w \vbf{I} - \sksym{\vbf{q}_v} \end{bmatrix},
\end{gather}
where $\circ$ denotes the quaternion multiplication operator.

Despite representing the same object, the use of quaternion and rotation matrix may be advantageous in some cases, and not in other cases (more explainations are given in Section \ref{sec: optimization on manifold}). In this paper we take liberty to use the rotation matrix interchangeably with its corresponding quaternion, as the conversion between them has been provided in the formulae \eqref{eq: quat to rot} and \eqref{eq: rot to quat}. 

\subsubsection{Special Euclidean Group} $\SE$ describes the group of rigid-body motions in 3D space whose definition is
\begin{equation*}
     \SE \triangleq \l\{\tf = \begin{bmatrix} \rot &\trans\\ \zero &1\end{bmatrix} \in \R^4 \colon \rot \in \SO, \trans \in \R^3\r\}.
\end{equation*}

Note that $\tf$ is often referred to as a \textit{transform}, $\rot$ as \textit{rotation} and $\trans$ as \textit{translation}. Similar to $\SO$, the group operation on $\SE$ is defined as $\tf_1 \circ \tf_2 = \tf_1\tf_2 \in \SE,\ \forall \tf_1, \tf_2 \in \SE$. Given a transform $\tf$ its inverse can be found as:
\begin{alignat*}{2}
    &\tf^{-1}
    &&= \begin{bmatrix} \rot^{-1} &-\rot^{-1}\trans\\ \zero\quad\quad &1\quad\quad\end{bmatrix}.
\end{alignat*}

In practice, a transform ${}^\fA_\fB\tf$ is used to convert a \textit{pose} in frame $\{\fB\}$ to $\{\fA\}$ via the operation $[{}^\fA\pos^\top,\ 1]^\top = {}^\fA_\fB\tf [{}^\fB\pos^\top,\ 1]^\top$, or more simply put, ${}^\fA\pos = {}^\fA_\fB\rot {}^\fB\pos + {}^\fA_\fB\trans$.

\subsection{Coordinate frames and Extrinsic Parameters} \label{sec: coordinate frames}

\begin{figure}[h]
    \centering
    \begin{overpic}[width=0.75\linewidth,
                    ]{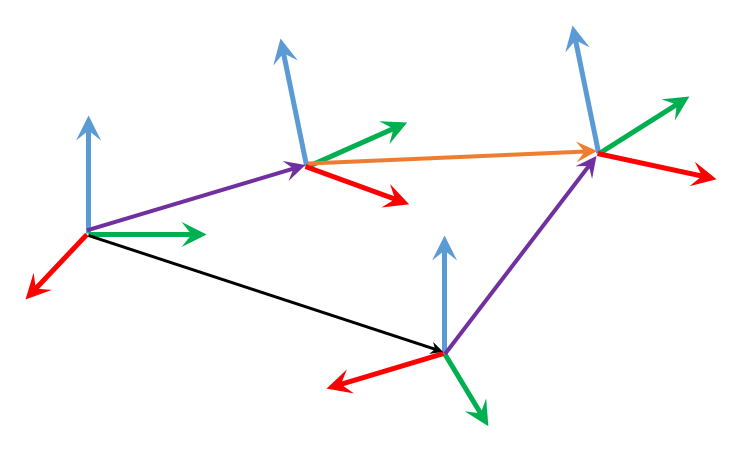}
		\put(00.50, 40.50){\footnotesize $\{\fW\}$}
		\put(40.00, 53.00){\begin{turn}{14} \footnotesize $\{\fB_k\} \equiv \{\fI_k\}$ \end{turn}}
		\put(80.50, 53.00){\begin{turn}{14} \footnotesize $\{\fC^i_k\}$ \end{turn}}
		\put(48.00, 25.00){\begin{turn}{00} \footnotesize $\{\fL^i\}$ \end{turn}}
		\put(23.00, 16.00){\begin{turn}{00} \footnotesize ${}^{\fW}_{\fL^i}\tf = ?$ \end{turn}}
		\put(20.00, 37.50){\begin{turn}{00} \footnotesize ${}^{\fW}_{\fB_k}\tf$ \end{turn}}
		\put(70.00, 20.00){\begin{turn}{00} \footnotesize ${}^{\fL^i}_{\fC^i_k}\tf$ \end{turn}}
		\put(57.50, 44.00){\begin{turn}{00} \footnotesize ${}^{\fB_k}_{\fC^i_k}\tf = \vbf{I}$ \end{turn}}
	\end{overpic}
	\caption{The coordinate frames that are considered under the sensor fusion scheme.}
	\label{fig: coordinate frames}
\end{figure}
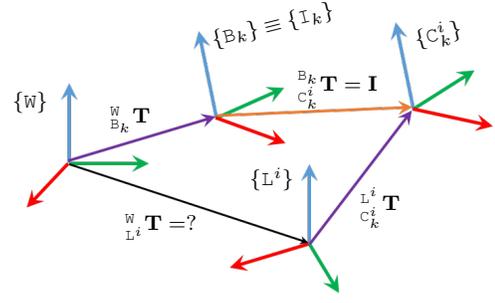

In reference to Fig. \ref{fig: coordinate frames}, we denote $\{\fB_k\}$ as the robot's \textit{body} coordinate frame at time $t_k$, and $\{\fW\}$ is the world frame. Obviously, the main goal of our work is to estimate the transform ${}^{\fW}_{\fB_k}\tf$:
\begin{equation*}
    {}^{\fW}_{\fB_k}\tf = \begin{bmatrix} {}^\fW_{\fB_k}\rot &{}^\fW_{\fB_k}\trans\\
                                      \zero        &1\end{bmatrix}
                    = \begin{bmatrix} \qtoR\l({}^\fW_{\fB_k}\quat\r) &{}^\fW_{\fB_k}\pos\\
                                      \zero                  &1\end{bmatrix},
\end{equation*}
where ${}^\fW_{\fB_k}\quat$ and ${}^\fW_{\fB_k}\pos$ are the orientation quaternion and position of the body frame \wrt the world frame at time $t_k$.


Following the common practice in the literature, we will consider the IMU-attached frame $\{\fI_k\}$ to coincide with the body frame, i.e. $\{\fI_k\} \equiv \{\fB_k\}, \forall k$.

Also in reference to Fig. \ref{fig: coordinate frames}, let us denote $\{\fC^i_k\}$ as a coordinate frame attached to the main sensor of the $i$-th OSL system, which could be a camera or a Lidar. In most practical OSL systems, we can obtain a measurement of the transform ${}^{\fL^i}_{\fC^i_k}\tf$, where $\{\fL^i\}$ is some local frame of reference, which usually coincides with $\{\fC^i_0\}$. 
Similar to the case with IMU, we can assume ${}^{\fB_k}_{\fC^i_k}\tf \equiv {}^{\fB}_{\fC^i}\tf, \forall k$ (i.e. the camera is rigidly mounted on the robot) and ${}^{\fB}_{\fC^i}\tf$ is known beforehand (in the literature ${}^{\fB_k}_{\fC^i_k}\tf$ is often called the \textit{extrinsic parameters} and can be measured or estimated from the OSL's data by some software tools). Indeed we can go a little bit further by assuming that ${}^{\fL^i_k}_{\fB_k}\tf = {}^{\fL^i_k}_{\fC^i_k}\tf \cdot \l({}^{\fB}_{\fC^i}\tf\r)^{-1}$ as the output of the OSL system. In effect, this is the same as assuming that all frames $\{\fC^i\}$ coincide with $\{\fB\}$, i.e. $ \forall \{\fC^i_k\} \colon {}^{\fB_k}_{\fC^i_k}\tf \triangleq \vbf{I}_{4\times4}.$
And henceforth, we will not have any further reference to the frame $\fC^i$ in this paper.
However, the transform ${}^\fW_{\fL^i}\tf$ remains a concern, as it is usually either unknown, very hard to obtain in practice, or subject to drift. In Section \ref{sec: vio factor}, based on the characteristics of OSL system, we will discuss converting this externally-referenced measurements into so-called \textit{pose displacement} measurements.


\subsection{Optimization problem on Manifold} \label{sec: optimization on manifold}

\subsubsection{Local reparameterization}

In robotic localization applications, we often encounter the following optimization problem
\begin{equation} \label{eq: main problem}
    \min_{\hat{\X} \in \M} f(\hat{\X}), 
\end{equation}
where $f(\cdot)$ is a cost function constructed from the sensor measurements, $\hat{\X}$ is the estimate of the robot states that may include position, orientation quaternion, velocity, etc., and $\M$ is a manifold where the robot states reside on.
Since the orientation states are often overparameterized (by nine variables for rotation matrices, and four for quaternion), the condition $\hat{\X} \in \M$ introduces extra nonlinear constraints to the problem, and solving such problem with both nonlinear cost function and constraints is not a trivial task.
Instead, the problem \eqref{eq: main problem} is often reparameterized so that the decision variable resides on the {linear} tangent space of the manifold defined at the current state $\hat{\X}$, i.e.
\begin{equation}
    \min_{\delta\hat{\X} \in \R^n} f\l(\Lift_{\hat{\X}}(\d\hat{\X})\r).
\end{equation}
where $\Lift_{\hat{\X}}(\cdot) \colon \R^n \to \M$ is called a local reparameterization at $\hat{\X}$, or \textit{retraction} function \cite{forster2016manifold}.
After the reparameterization, for each iteration in the Gauss-Newton method, we can calculate the optimal gradient $\d\hat{\X}^\star$ and then "retract" the solution $\d\hat{\X}^\star$ from the tangent space back to the manifold using the operation $\hat{\X} \leftarrow \Lift_{\hat{\X}}(\d\hat{\X}^\star)$.
In practice, the mappings \eqref{eq: rodrigues formula} and \eqref{eq: vec to quat} are used to define the local parameterization of rotations as follows:
\begin{equation}
    \Lift_{\hat{\rot}}(\d\eulvec) = \hat{\rot} \Exp(\d\eulvec),\qquad \Lift_{\hat{\quat}}(\d\eulvec) = \hat{\quat}\circ\calE(\d\eulvec).\ 
\end{equation}
For a vector state such as position and velocity, the retraction function is simply defined as:
\begin{equation}
    \Lift_{\hat{\pos}}(\d\pos) = \hat{\pos} + \d\pos,\qquad \Lift_{\hat{\vel}}(\d\vel) = \hat{\vel} + \d\vel.\ 
\end{equation}

Note that in \eqref{eq: rodrigues formula} and \eqref{eq: vec to quat}, the use of trigonometric function can be computationally expensive, and the variable at the denominator has the potential of causing division by zero fault. This is why we often employ the following approximations
\begin{equation}\label{eq: d quat approx}
    \Lift_{\hat{\rot}}(\d\eulvec)  \simeq \rot(\ident + \sksym{\d\eulvec}),\quad 
    \Lift_{\hat{\quat}}(\d\eulvec) \simeq \quat \circ \begin{bmatrix}
                                                    1\\
                                                 \frac{1}{2}\d\eulvec
                                                 \end{bmatrix},
\end{equation}
which only involve basic arithmetic operations. As can be seen in later parts, the former is convenient for calculating the Jacobians and the covariance, while the latter is more convenient in calculating the rotational residual and the retraction step.

\subsubsection{Jacobian and the chain rule}

One other important concept for the optimization on manifold is the Jacobian of a function $f(\X) \colon \M \to \mathcal{N}$ acting on manifolds (e.g. $f(\rot) = \rot \vbf{y}$, where $\vbf{y} \in \R^3$, $\M = \SO$ and $\mathcal{N} = \R^3$), the Jacobian (or more precisely \textit{right Jacobian}) of $f(\X)$ is defined as
\begin{equation} \label{eq: Jacobian for manifold}
    \Jcb^{f(\X)}_{\X} \triangleq \frac{\partial\Log_{\mathcal{N}}\l[f(\X)^{-1} \circ f(\Lift_\X(\pith))\r]}{\partial\pith}\Bigg|_{\pith = 0},
\end{equation}
where $\Log_{\mathcal{N}}(\X)$ is the mapping from the manifold $\mathcal{N}$ to the \textit{tangent space at the identity} of $\mathcal{N}$. Recall that in \eqref{eq: right jacobian definition}, we have applied \eqref{eq: Jacobian for manifold} in the case where $f(\X) = \Exp(\X)$, $\M = \R^3$ and $\mathcal{N} = \SO$.

From  \cite{sola2018micro} the chain rule simply states that for $\mathcal{Y} = f(\X)$ and $\mathcal{Z} = g(\mathcal{Y})$, we have
\begin{equation}
    \Jcb^{\mathcal{Z}}_{\mathcal{X}} = \Jcb^{\mathcal{Z}}_{\mathcal{Y}} \Jcb^{\mathcal{Y}}_{\mathcal{X}},
\end{equation}
which can be generalized to any number of variables.

\section{Methodology} \label{sec: problem description}

\subsection{Problem formulation}

Let us denote all of the states to be estimated at time $t_k$ as $\X_k$ as follows:
\begin{align}
    \X_k &= \Big( \quat_k,\ \pos_k,\ \vel_k,\ \bias^{\o}_{k},\ \bias^{a}_{k}\Big), \label{eq: state vector}
\end{align}
where $\quat_k = {}^\fW_{\fB_k}\quat$, $\pos_k = {}^\fW_{\fB_k}\pos$, $\vel_k = \dot{\pos}_k \in \R^3$ are respectively the robot's orientation quaternion, position and velocity \wrt the world frame $\{\fW\}$ at time $t_k$; $\bias^{a}_{k},\ \bias^{\o}_{k} \in \R^3$ are respectively the IMU accelerometer and gyroscope biases.

Hence, we define the state estimate at each time step $k$, and the sliding windows as follows:
\begin{alignat}{2}
    &\hat{\X} &&= \l(\hat{\X}_k,\ \hat{\X}_{k+1} \dots\ \hat{\X}_{k+M}\r), \label{eq: X hat k to k+M}\\
    &\hat{\X}_k &&= \Big( \hat{\quat}_k,\ \hat{\pos}_k,\ \hat{\vel}_k,\ \hat{\bias}^{\o}_{k},\ \hat{\bias}^{a}_{k}\Big), \label{eq: X hat k}
\end{alignat}
where $M \in \N$ is the size of the sliding windows.


\begin{figure}[t]
    \centering
    \begin{overpic}[width=\linewidth,
                        ]{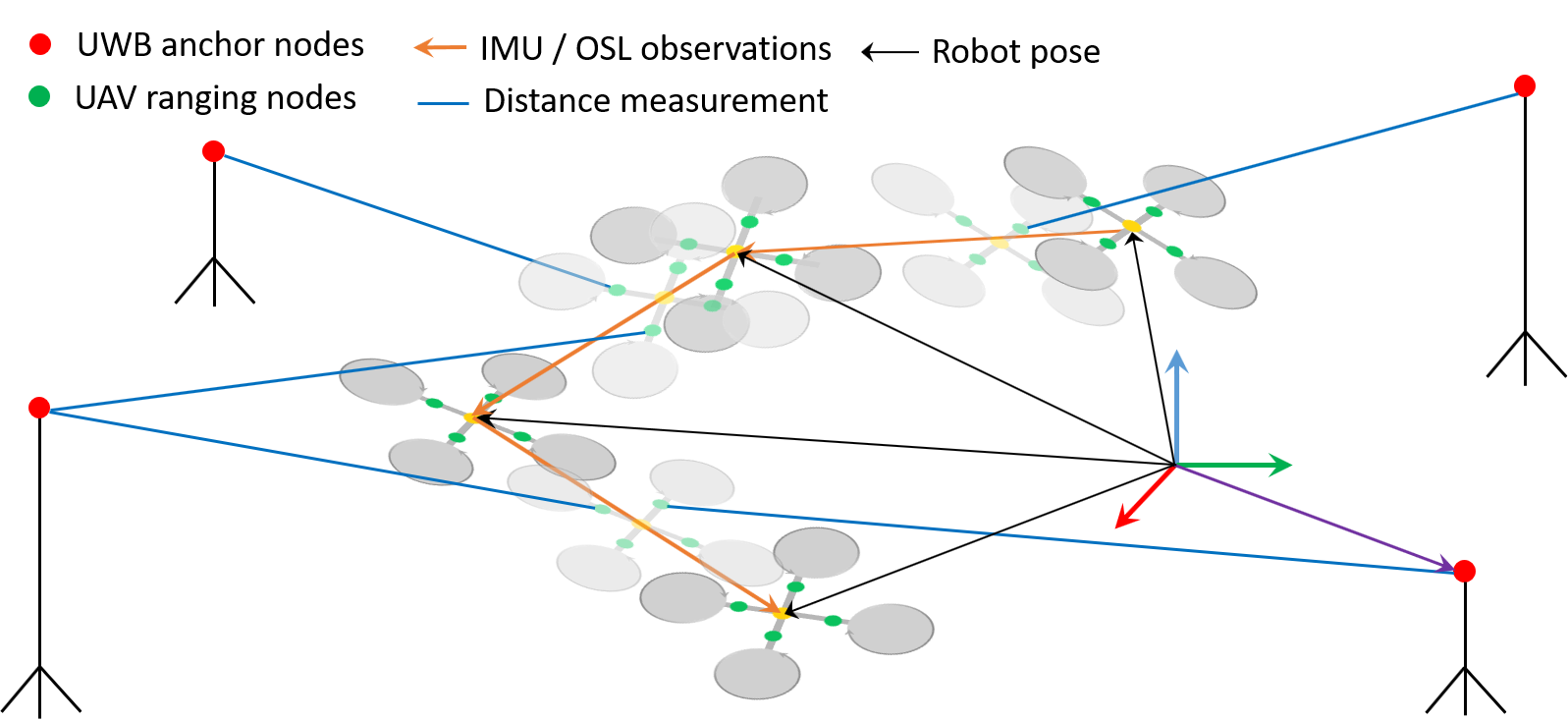}
		\put(76.50, 18.50){\footnotesize $\{\fW\}$}
        \put(74.00, 24.50){\footnotesize $\X_k$}
        \put(60.00, 25.00){\footnotesize $\X_{k+1}$}
        \put(50.00, 20.00){\footnotesize $\X_{k+2}$}
        \put(55.50, 13.50){\footnotesize $\X_{k+3}$}
        \definecolor{distancolor}{RGB}{0, 112, 192}
        \put(79.00, 39.00){\footnotesize \textcolor{distancolor}{$\breve{d}_{k+1}^1$}}
        \put(20.00, 25.00){\footnotesize \textcolor{distancolor}{$\breve{d}_{k+2}^2$}}
        \put(26.00, 34.00){\footnotesize \textcolor{distancolor}{$\breve{d}_{k+2}^1$}}
        \put(70.00, 7.00){\footnotesize \textcolor{distancolor}{$\breve{d}_{k+3}^2$}}
        \put(14.00, 14.00){\footnotesize \textcolor{distancolor}{$\breve{d}_{k+3}^1$}}
        \definecolor{anchorcolor}{RGB}{112, 48, 160}
        \put(84.00, 13.00){\footnotesize \textcolor{anchorcolor}{${}^\fW\vbf{x}_{k+3}^i$}}
	\end{overpic}
	\caption{Illustration of a sliding window consisting of four consecutive states $\X_k,\dots\X_{k+3}$ of the robot registered at for time steps $t_k, \dots t_{k+3}$. For a closer look, during the interval $(t_{k+1}, t_{k+2}]$, we have two body-offset UWB distances $\breve{d}_{k+2}^1$ and $\breve{d}_{k+2}^2$, hence the subscript $k+2$ and the superscripts $1$ and $2$ are used. Between two consecutive poses, IMU and OSL sensors can provide relative displacement observation. Hence UWB, IMU and OSL data will be fused together to estimate the robot state $\X_k$.
}
	\label{fig: overview}
\end{figure}
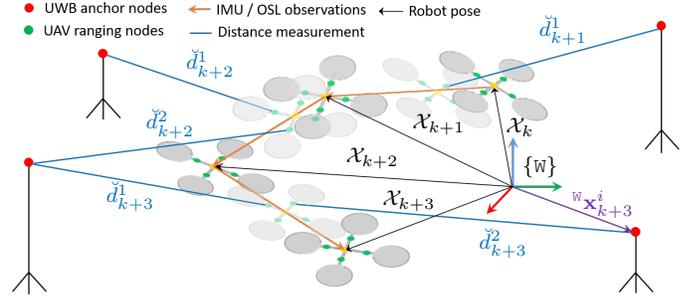

Thus, we will construct a cost function based on the observations from UWB, IMU and OSL measurements at every new time step. As can be seen in Fig. \ref{fig: overview}, UWB, IMU and OSL will provide observations that couple two consecutive states in the sliding window.
Using all of the observation obtained from UWB, OSL and IMU, a cost function $f(\hat{\X})$ can be constructed at each time step $t_{k+M}$ as:
\begin{align}
    f(\hat{\X}) &\triangleq
    \Bigg\{\mkern13mu \sum_{m = k}^{k+M-1} \sum_{i = 1}^{N_{\U}^{m}} \norm{\resi_{\U}(\hat{\X}_{m}, \hat{\X}_{m+1}, \U_m^i)}^2_{\vbf{P}_{\U^i}^{-1}} \nonumber\\
    &\qquad+ \sum_{m = k}^{k+M-1} \sum_{i = 1}^{N_{\V}^m} \norm{\resi_\V(\hat{\X}_{m}, \hat{\X}_{m+1}, \V_m^i)}^2_{\vbf{P}_{\V^i}^{-1}} \nonumber\\
    &\qquad+ \sum_{m = k}^{k+M-1} \norm{\resi_\I(\hat{\X}_{m}, \hat{\X}_{m+1}, \I_m)}^2_{\vbf{P}_{\I_m}^{-1}}\quad \Bigg\}, \label{eq: cost function}
\end{align}
where $\resi_{\U}(\cdot)$, $\resi_\V(\cdot)$, $\resi_\I(\cdot)$ are the \textit{residuals} constructed from UWB, IMU and OSL observations; $\vbf{P}_{\U^i}$, $\vbf{P}_{\V^i}$, $\vbf{P}_{\I_k}$ are the covariance matrices of the measurement error, and $N_{\U}^k$, $N_{\V}^k$ respectively are the number of UWB and OSL observations obtained in the inverval $[t_{k},\ t_{k+1}]$.

Note that each term $\norm{r(\cdot)}^2_{\vbf{P}^{-1}}$ above is also referred to as a \textit{cost factor} or simply \textit{factor}. In Section \ref{sec: cost factors} we will elaborate on the construction of these terms.

\subsection{Workflow}

\begin{figure}
    \centering
    \begin{overpic}[width=\linewidth,
                        ]{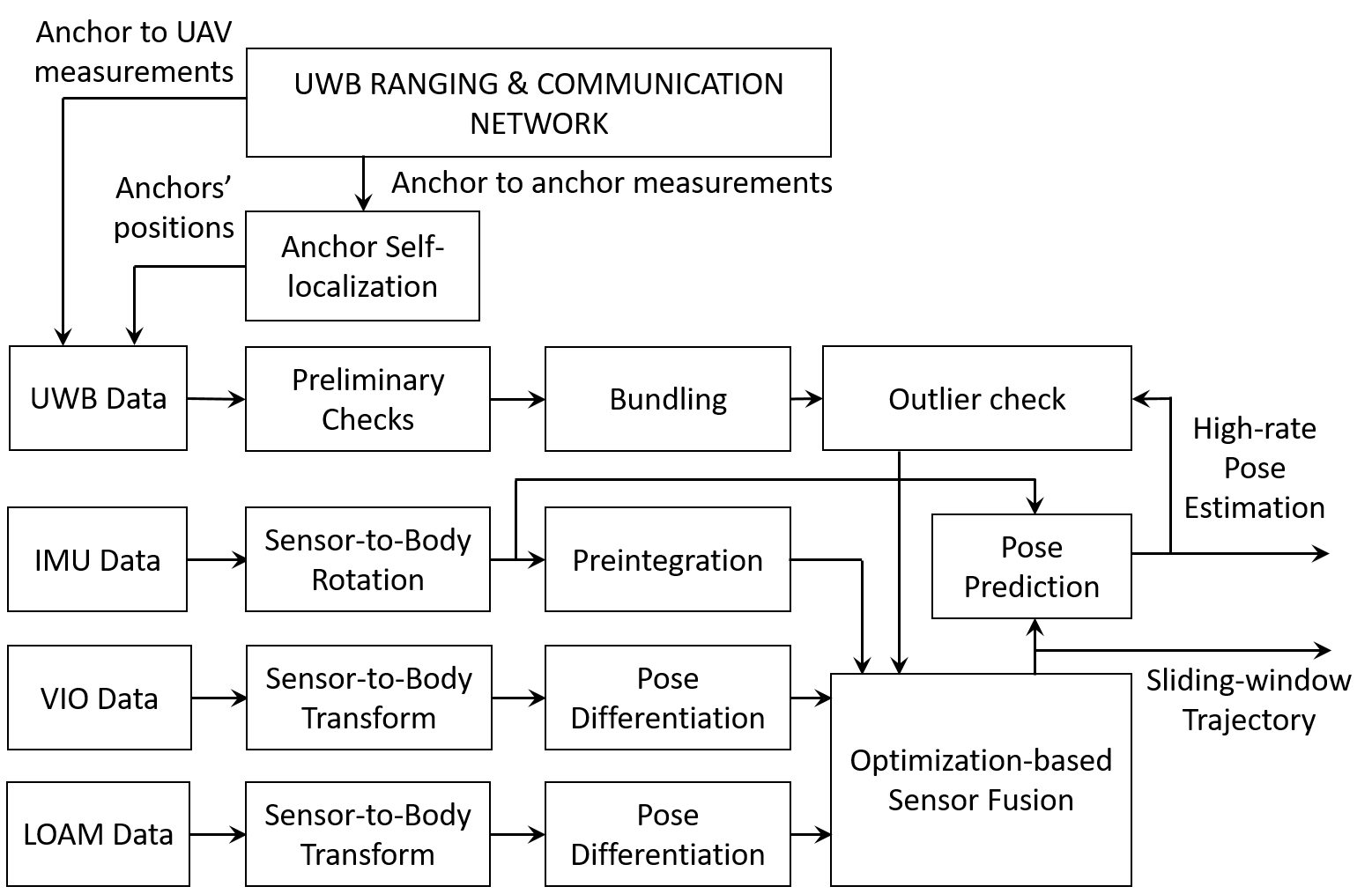}
	\end{overpic}
	\caption{Overview of the system's main components.}
	\label{fig: block diagram}
\end{figure}

Fig. \ref{fig: block diagram} provides an overview of our localization system. As having introduced in Section \ref{sec: intro}, the UWB ranging and communication network (Fig. \ref{fig: block diagram}, top part) is employed to gather the distance between the anchors, after they have been deployed in the field. These distances will be then used to calculate the position of the anchors by using a self-localization algorithm similar to our previous works \cite{nguyen2016ultra}. Once the anchors' positions have been calculated, they will be associated with the UWB measurements. Thus, a UWB observation consists of three main pieces of information, first is the position of the UWB node in the UAV's body frame, second is the UWB anchor position in the chosen world frame, and third is the distance between these the UWB node and the UWB anchor. Note that some information regarding the sampling times will also be used, as explained Section \ref{sec: range factors}.

After the anchor self-localization process, UWB, IMU and OSL data will go through multiple stages before they are fused together in the Optimization-based Sensor Fusion (OSF) process.

In the first stage, for each UWB measurement obtained, it will go through some preliminary checks to eliminate bad measurements. These checks include comparing the measurement's signal over noise ratio (SNR), leading-edge-detection quality, and rate-of-change with some user-defined thresholds, etc. Only those that are within these thresholds will be passed to the buffer for subsequent stages.
For IMU and OSL, we will transform these measurements into the body frame using the fixed extrinsic transforms (see Section \ref{sec: coordinate frames}). These transformed measurements are then passed to the corresponding buffers for the next stages.

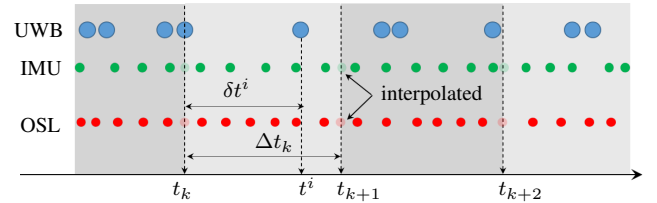
\begin{figure}
    \centering
    \begin{overpic}[width=0.95\linewidth,
                        ]{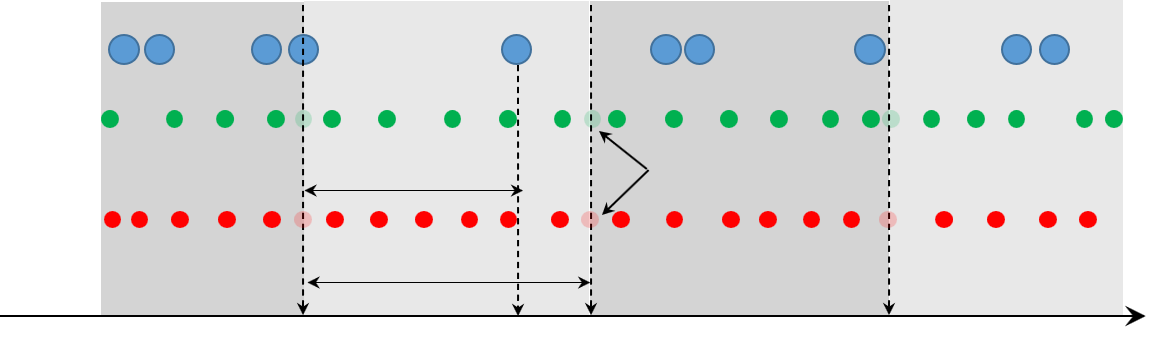}
                        \put(-1.0,  23.5){\footnotesize UWB}
                        \put(0.0,   17.5){\footnotesize IMU}
                        \put(0.0,   8.5){\footnotesize OSL}
                        \put(57.0,  13.5){\footnotesize interpolated}
                        \put(32.0,  14.0){\footnotesize $\dt^i$}
                        \put(37.0,  6.0){\footnotesize $\Dt_k$}
                        \put(24.0, -1.5){\footnotesize $t_{k}$}
                        \put(44.0, -1.5){\footnotesize $t^i$}
                        \put(50.0, -1.5){\footnotesize $t_{k+1}$}
                        \put(75.5, -1.5){\footnotesize $t_{k+2}$}
	\end{overpic}
	\caption{Synchronization scheme:
	at every time step $t_k$, we would be able to obtain a set of UWB observations (see \eqref{eq: distance model}), an IMU preintegration observation (see \eqref{eq: pIMU measurement}), and a pose-displacement measurement from each OSL system (see \eqref{eq: pose displacement measurement model}). Note that in practice multiple UWB measurements can be acquired simultaneously, hence there will be no other data from other sensors between them (more details in Sec. \ref{sec: experiment}), or that some measurements can be lost in some intervals. This will cause some significant challenges that are discussed in more details in Remark \ref{rem: sync issue}.
	}
	\label{fig: sync scheme} 
\end{figure}

In the second stage, after every fixed period of time, we extract UWB measurements together with the IMU and OSL measurements from the buffers to construct some observations,
namely IMU preintegration $\I_k$, OSL-based pose-displacement $\V^i_k$ and UWB body-offset ranges $\U^i_k$.
In reference to Fig. \ref{fig: sync scheme}, these measurements are extracted at the time instance $t_{k+1}$, which is also used to register a new state estimate $\hat{\X}_{k+1}$.
Note when the new step $t_{k+1}$ is created, we can use it to make an interpolated IMU and OSL measurements from the immediate preceding and succeeding samples (see Fig. \ref{fig: sync scheme}). Hence all of the IMU measurements within $[t_k,\ t_{k+1}]$, including the interpolated values, will be used to calculate the preintegration measurements, and the initial and final OSL values in the interval will be used to calculate the pose displacement measurements. For UWB, each observation also includes the time offset $\dt^i$ from the latest earlier time step (see Fig. \ref{fig: sync scheme}), which is needed for the construction of the range factor in Section \ref{sec: range factors}.

In the final stage, the stack of UWB data will be subjected to another check to see if any measurement is an outlier. This is done by comparing the measurement with the predicted range value using the IMU-predicted robot pose at time $t_{k} + \dt^i$. When the UWB measurements have passed the outlier check, we can finally use them to construct the cost function, together with the preintegrated IMU measurements and the OSL-derived pose displacement measurements. The function will be then optimized to provide an estimate of the robot states in the sliding windows. The latest state estimate in the sliding windows will be combined with the latest IMU measurements in the buffer, which are leftovers from the last extraction plus the newly arrived samples during the optimization process, to calculate the high-rate pose estimate. This high-rate pose estimate is also used to perform the outlier check on the UWB measurements that was mentioned earlier.

After the optimization has been completed, the oldest state estimate and its associated measurements will be abandoned, and a new time step will be registered when the conditions mentioned earlier are met.

\begin{rem} \label{rem: sync issue}

The design of the synchronization scheme described above generalizes our method compared to previous works \cite{fang2018model, fang2019graph, wang2017ultra, nguyen2019loosely, nguyen2019tightly, nguyen2020tightly, cao2020vir} and is a necessary requirement when fusing such an extensive set of sensor data. In \cite{fang2018model, fang2019graph, wang2017ultra, nguyen2019loosely}, the creation of a new state $\hat{\X}_k$ is synchronized with the acquisition of a new UWB sample, hence when multiple UWB samples are obtained at the same time, or UWB ranging is temporarily lost, the estimation scheme can be destabilized. Moreover, when UWB rate increases, real-time processing may not be guaranteed, since the number of UWB factor increases (hence increasing the computational complexity), but the interval $\Dt_k$ available for the optimization process is reduced. On the other hand, in \cite{nguyen2020tightly, cao2020vir}, the creation of new state is synchronized with the camera sample times, and only UWB samples that are closest to the camera sample times are used, while others are discarded. In our framework, the time steps $t_k$ are independent of the timing of the sensor data. Hence when more sensors are integrated and complexity increases, we can easily adjusted $\Dt_k$, or even skip the optimization in one step (see Section \ref{sec: init and realtime}) to ensure real-time estimation. 
\end{rem}

\subsection{Initialization and real-time optimization} \label{sec: init and realtime}

While many OSL based methods require an elaborate initialization motions before actual operation \cite{forster2014svo, murorb2mono}, this appears to be a very difficult task when working with UAV platforms or medium size or beyond. Thus, in this work, initialization is done by collecting about 100 UWB measurements on startup, then run the optimization with several initial yaw values in $[0, 2\pi)$, and the estimate that yields the smallest cost is chosen. Note that roll and pitch are initialized directly from the angle between the measured gravity vector and the body frame's z axis.

During the estimation process, sometimes the optimization operation can return quickly before the next batch of measurements are admitted to the sliding windows. In this case, we can try to "explore" other minima of the cost function by nudging the yaw value around the current estimate, and rerun the optimization. Whenever a smaller cost is obtained, the estimate will be updated to this new value. Moreover, sometimes the optimization may elapse when multiple batches of measurements are still in the queue. In this case we can skip the optimization for a few steps and slide the window forward to use the latest patch for constructing the cost function, so as to ensure the real-time operation of the algorithm.

\section{Construction of the cost factors} \label{sec: cost factors}

In this section, we will derive the factor for each type of observation in the cost function \eqref{eq: cost function}. In general, the basic procedure is as follows: 1) we construct the observation model that states the relationship of the sensor observation, the robot true states, and the measurement error; 2) we calculate the covariance matrix of the error in order to properly apply the weightage of each least-square term in the optimization problem; 3) we calculate the residual (difference) between each observation and its predicted value, which is obtained by plugging the current state estimate into the mathematical observation model; 4) we calculate the Jacobian for the fast implementation of the optimization process.


\subsection{OSL factors} \label{sec: vio factor}

In this section, we discuss the construction of pose-displacement observation from OSL data. Note that we would omit the OSL system index $i$ in \eqref{eq: cost function} since the treatment is the same for all OSL systems available on the UAV.

\subsubsection{Observation model} OSL systems can have slightly different implementations (monocular or stereo, RGBD or Lidar, etc.) and characteristics (the definition of features, overall accuracy and data rate, etc.). However, as having discussed in Section \ref{sec: coordinate frames}, due to such diversity, in this paper we consider OSL systems working in a manner similar to odometer, where they essentially measure the \textit{relative motion} of the sensor between two time instances, and then integrate these relative measurements to provide an estimate of the robot pose \wrt some local frame of reference \{\fL\}. To this end, we model the OSL measurement as follows.
\begin{subequations} \label{eq: vio measurements}
    \begin{align}[left = \empheqlbrace]
      {}^{\fL}_{\fB_{k+1}}\breve{\quat}
      &= {}^{\fL}_{\fB_{k}}\breve{\quat} \circ \l(\D\quat_k \circ \calE(\eulvec_k)\r),
      \label{eq: vio quat measurements}\\
      {}^{\fL}_{\fB_{k+1}}\breve{\pos}
      &=
      {}^{\fL}_{\fB_k}\breve{\pos} + {}^{\fL}_{\fB_k}\breve{\rot}\l(\D\pos_k + \vbf{d}_k\r),
      \label{eq: vio pos measurements}
    \end{align}
\end{subequations}
where $\calE(\eulvec_k)$, $\vbf{d}_k$ are some errors in the estimation process, $\D\quat_k$ and $\D\pos_k$ are the relative rotation and displacement from time $t_k$ to $t_{k+1}$, which are formally defined as:
\begin{subequations} 
    \begin{align}[left = \empheqlbrace]
      \D\quat_k
      &\triangleq {}^{\fW}_{\fB_k}\quat^{-1} \circ {}^{\fW}_{\fB_{k+1}}\quat,
      \label{eq: vio relative rotation}\\
      \D\pos_k
      &\triangleq {}^{\fL}_{\fB_k}\rot^{-1}\l({}^{\fL}_{\fB_{k+1}}\pos - {}^{\fL}_{\fB_{k}}\pos\r). \label{eq: vio relative desplacement}
    \end{align}
\end{subequations}

From the user's end, only ${}^{\fL}_{\fB_{k+1}}\breve{\quat}$ and ${}^{\fL}_{\fB_{k+1}}\breve{\pos}$ are accessible. Thus, given these measurements, we can retrieve the pose-displacement observation as follows:
\begin{subequations} \label{eq: pose displacement measurement model}
    \begin{alignat}{2}[left = \empheqlbrace]
    &\V_{k} &&\triangleq\l(\D\breve{\quat}_k,\ \D\breve{\pos}_k\r),
    \\
    &\D\breve{\quat}_k &&\triangleq \breve{\quat}_k^{-1} \circ \breve{\quat}_{k+1} = \D \quat_k \circ \calE(\eulvec_k)  ,
    \label{eq: quat displacement measurement model}
    \\
    &\D\breve{\pos}_k  &&\triangleq \breve{\rot}_k^{-1}\l(\breve{\pos}_{k+1} - \breve{\pos}_k\r) = \D \pos_k + \vbf{d}_k.
    \label{eq: pos displacement measurement model}
    \end{alignat}
\end{subequations}

\subsubsection{Observation error covariance}

In reference to \eqref{eq: pose displacement measurement model}, we model the ``noises" $\eulvec_k$ and $\vbf{d}_k$ as independent zero-mean Gaussians with the following covariance
\begin{equation}
    \vbf{P}_{\V_k} = \cov\l(\begin{bmatrix} \eulvec_k\\ \vbf{d}_k\end{bmatrix}\r) = \Dt_k\mathrm{diag}(\s_{\V_1}^2,\ \s_{\V_2}^2,\ \dots\ \s_{\V_6}^2).
\end{equation}
where $\s_{\V_1}^2,\ \s_{\V_2}^2,\ \dots\ \s_{\V_6}^2$ are some parameters that can be empirically determined.

\subsubsection{Residuals} Given the observations in \eqref{eq: pose displacement measurement model}, we can calculate the residuals from OSL as follows:
\begin{equation} \label{eq: OSL residual}
    \resi_{\V}(\hat{\X}_k, \hat{\X}_{k+1}, \V_{k})
        \triangleq \begin{bmatrix}
                        \resi_{\D\breve{\quat}}\\
                        \resi_{\D\breve{\pos}}
                    \end{bmatrix}
        = \begin{bmatrix}
            2\vbf{vec}(\D\breve{\quat}_k^{-1} \circ \D\hat{\quat}_k)\\
            \D\hat{\pos}_k - \D\breve{\pos}_k
           \end{bmatrix},
\end{equation}
where $\vbf{vec}(\cdot)$ returns the vector part of the quaternion, $\D\hat{\quat}_k$ and $\D\hat{\pos}_k$ are the estimated pose displacement at time $t_k$, which can be calculated in the same manner with \eqref{eq: pose displacement measurement model}:
\begin{align}
    \D\hat{\quat}_k \triangleq \hat{\quat}_{k}^{-1} \circ \hat{\quat}_{k+1},\quad \D\hat{\pos}_k  \triangleq  \hat{\rot}_k^{-1}\l(\hat{\pos}_{k+1} - \hat{\pos}_k\r). \label{eq: Delta qhat}
\end{align}

\subsubsection{Jacobian}

Let us first examine $\Jcb^{\resi_{\D\breve{\quat}}}_{\hat{\quat}_k}$. From \eqref{eq: OSL residual} and \eqref{eq: Delta qhat} and the fact that $\D\breve{\quat}_k^{-1} \circ \hat{\quat}_k^{-1}\circ\hat{\quat}_{k+1} = (\hat{\quat}_{k+1}^{-1}\circ\hat{\quat}_{k} \circ \D\breve{\quat}_k)^{-1}$, we have $\resi_{\D\breve{\quat}} = -2\vbf{vec}\l[\hat{\quat}_{k+1}^{-1}\circ\hat{\quat}_{k} \circ \D\breve{\quat}_k\r]$ since $\vbf{vec}(\quat) = -\vbf{vec}(\quat^{-1})$ over a unit quaternion $\quat$. Hence:
\begin{align*}
    \resi_{\D\breve{\quat}}(\Lift_{\hat{\quat}_k}(\d\pith_k))
     &\simeq
    -2\vbf{vec}\l[\hat{\quat}_{k+1}^{-1}
                    \circ
                  \hat{\quat}_{k}
                    \circ
                  \begin{bmatrix}1\\ \frac{1}{2}\d\pith_k\end{bmatrix}
                    \circ
                  \D\breve{\quat}_k
               \r]\\
    &= -2\vbf{vec}\l[\qmatR{\D\breve{\quat}_k}
                   \qmatL{\hat{\quat}_{k+1}^{-1}
                     \circ
                   \hat{\quat}_{k}}
                   \begin{bmatrix}1\\ \frac{1}{2}\d\pith_k\end{bmatrix}
                  \r],
    \end{align*}
and we can straightforwardly obtain
\begin{equation} \label{eq: jcb rdq}
    \Jcb^{\resi_{\D\breve{\quat}}}_{\hat{\quat}_k} = -\brc_3\l(\qmatR{\D\breve{\quat}_k} \qmatL{\hat{\quat}_{k+1}^{-1} \circ \hat{\quat}_{k}}\r).
\end{equation}

Similarly we can obtain the Jacobian $\Jcb^{\resi_{\D\breve{\quat}}}_{\hat{\quat}_k}$ as follows
\begin{equation*}
    \Jcb^{\resi_{\D\breve{\quat}}}_{\hat{\quat}_{k+1}} = \brc_3\l(\qmatL{\D\breve{\quat}_k^{-1} \circ \hat{\quat}_{k+1}^{-1} \circ \hat{\quat}_{k}}\r).
\end{equation*}

To obtain $\Jcb^{\resi_{\D\breve{\pos}}}_{\hat{\quat}_k}$, by employing \eqref{eq: d quat approx}, \eqref{eq: skew and cross}, we can obtain
\begin{align*}
    &\resi_{\D \breve{\pos}}(\Lift_{\hat{\quat}_k}(\d \pith_k) )
    \simeq (\ident + \sksym{\d\pith_k})^\top\hat{\rot}_k^\top(\hat{\pos}_{k+1} - \hat{\pos}_{k})\\
    &\qquad= -\sksym{\d\pith_k}\hat{\rot}_k^\top(\hat{\pos}_{k+1} - \hat{\pos}_{k}) + \hat{\rot}_k^\top(\hat{\pos}_{k+1} - \hat{\pos}_{k})\\
    &\qquad= \quad \sksym{\hat{\rot}_k^\top(\hat{\pos}_{k+1} - \hat{\pos}_{k})}\d\pith_k + \hat{\rot}_k^\top(\hat{\pos}_{k+1} - \hat{\pos}_{k}).
\end{align*}
Thus, taking the derivative over $\d\pith_k$ yields
\begin{equation*}
    \Jcb^{\resi_{\D\breve{\pos}}}_{\hat{\quat}_k} \simeq \sksym{\hat{\rot}_k^\top(\hat{\pos}_{k+1} - \hat{\pos}_{k})}.
\end{equation*}

Following the same steps, we can easily find that
\begin{equation}
    \Jcb_{\hat{\pos}_{k}}^{\resi_{\D\breve{\pos}}} \simeq -\hat{\rot}_k^\top,\quad \Jcb_{\hat{\pos}_{k+1}}^{\resi_{\D\breve{\pos}}} \simeq \hat{\rot}_k^\top.
\end{equation}

\subsection{IMU preintegration factors:}

\subsubsection{IMU kinematic models}
Before exploring the preintegration observation models, let us recall some features of IMU.
First, let us notice the following kinematic model of the robot pose and velocity \wrt to the world frame:
\begin{align}
    \dot{\quat}_t =  {\quat}_t \circ \begin{bmatrix}0\\ \angvel_t\end{bmatrix},\quad
    \
    \dot{\pos}_t = \vel_t,\quad
    \
    \dot{\vel}_t = \rot_t \accel_t,\label{eq: imu kine model}
\end{align}
where $\angvel_t$, $\accel_t$ are the angular velocity and acceleration of the robot measured in the body frame. In reality we only have access to IMU measurements that are corrupted by noises and biases, i.e.:
\begin{align}
    &\breve{\angvel}_t = {\angvel}_t + \bias^{\o}_{t} + \noise^{\o}_{t}, \label{eq: angvel imu}\\
    &\breve{\accel}_t  = {\accel}_t  + \rot_{t}^\top \grav + \bias^{a}_{t} + \noise^{a}_t, \label{eq: accel imu}\\
    &\dot{\bias}^{\o}_{t} = \noise^{b\o}_{t},\ \dot{\bias}^{a}_{t}  = \noise^{ba}_{t},
\end{align}
where $\grav = [0,\ 0,\ g]^\top$ is the gravitational acceleration, $\bias^{\o}_{t}, \bias^{a}_{t}$ are respectively the biases of the gyroscope and the accelerometer, and $\noise^{\o}_{t}$, $\noise^{a}_t$, $\noise^{b\o}_{t}$, $\noise^{ba}_{t}$ are some zero-mean Gaussian noises, whose standard deviations are $\s_{\vbf{\eta}_\o}$, $\s_{\vbf{\eta}_a}$, $\s_{\vbf{\eta}_{b\o}}$, $\s_{\vbf{\eta}_{ba}}$, respectively.


\subsubsection{IMU preintegration}

Let us substitute \eqref{eq: angvel imu} and \eqref{eq: accel imu} to the kinematic model \eqref{eq: imu kine model}, \eqref{eq: imu kine model}, \eqref{eq: imu kine model}, and rewrite them in the integral form as follows:
\begin{align}
    {\quat}_{k+1}
    %
    &= {\quat}_{k} \circ \Mint_{t_k}^{t_{k+1}} \frac{1}{2}\Omg(\breve{\angvel}_s - \bias_s^\o - \noise_s^\o) {}\quat_s\label{eq: quat int},\\
    \pos_{k+1}  &= \pos_k + \vel_k\Dt_k - \frac{1}{2}\grav\Dt^2_k\nonumber\\
                &\quad+ {}^\fW\rot_k\Int{t_k}{t_{k+1}}\Int{t_k}{u} {}^{k}\rot_{s}\l( \breve{\accel}_s - \bias_s^a - \noise_s^a \r) ds\ du,\label{eq: pos int}\\
    \vel_{k+1}  &= \vel_k - \grav\Dt_k \\
                &\quad+ {}^\fW\rot_k\Int{t_k}{t_{k+1}} {}^{k}\rot_{s}\l( \breve{\accel}_s - \bias_s^a - \noise_s^a\r) ds, \label{eq: vel int}
\end{align}
where $\Mint(\cdot)$ denotes the integration operator on manifold, and a few short-hands have been employed
\begin{align}
    &{}^{k}{\quat}_{s} \triangleq {}^{\fB_k}_{\fB_s}{\quat},
    \qquad 
    {}^\fW\rot_{k} = \rot_k,
    \qquad 
    {}^{k}\rot_{s}  \triangleq {}^{\fB_k}_{\fB_s}\rot\\
    &\Omg(\angvel) \triangleq \begin{bmatrix} 0      &-{\bm\o}^\top\\
                                              {\bm\o} &-\sksym{\bm\o}
                               \end{bmatrix},
    \quad \Dt_k \triangleq t_{k+1} - t_k.
\end{align}

We can reorganize the terms in \eqref{eq: quat int}, \eqref{eq: pos int}, \eqref{eq: vel int} to reveal the observation model on the robot states. Specifically,
\begin{align}
    \pig_{k+1} &\triangleq \Mint_{t_k}^{t_{k+1}}\frac{1}{2}\Omg(\breve{\angvel}_s - \bias_s^\o - \noise_s^\o){}\quat_s
    = \quat_{k}^{-1} \circ \quat_{k+1}, \label{eq: preint gamma ideal}\\
    \pib_{k+1} &\triangleq \Int{t_k}{t_{k+1}} {}^{k}\rot_{s} (\breve{\accel}_s - \bias_s^a - \noise_s^a ) ds\nonumber\\
               &= \rot_k^{-1}\l(\vel_{k+1} - \vel_k + \grav\Dt_k\r),\label{eq: preint beta ideal}\\
    \pia_{k+1} &\triangleq \Int{t_k}{t_{k+1}}\Int{t_k}{u} {}^{k}\rot_{s} ( \breve{\accel}_s - \bias_s^a - \noise_s^a ) ds\ du\nonumber\\
                   &= \rot_k^{-1}(\pos_{k+1} - \pos_k - \vel_k\Dt_k + \frac{1}{2}\grav\Dt^2_k).\label{eq: preint alpha ideal}
\end{align}

Thus, we can see from \eqref{eq: preint gamma ideal}, \eqref{eq: preint beta ideal}, \eqref{eq: preint alpha ideal} that the integrals obtained from the IMU measurements, seen on the left hand side, are functions of the robot states on the right hand side.

\subsubsection{Observation model}

In reality we do not know the exact value of the IMU noises and bias. However, if we ignore the noise and use some nominal values $\bar{\bias}^{a}_{k}$, $\bar{\bias}^{\o}_{k}$ for the biases during $[t_k, t_{k+1}]$, we can calculate the following tuple of preintegrated IMU observations as follows:
\begin{equation}\label{eq: pIMU measurement}
    \I_k = (\breve{\pia}_{k+1}, \breve{\pib}_{k+1}, \breve{\pig}_{k+1}),
\end{equation}
where
\begin{align}
    \breve{\pia}_{k+1} &\triangleq \Int{t_k}{t_{k+1}}\Int{t_k}{u} {}^{k}\breve{\rot}_{s} (\breve{\accel}_s - \bar{\bias}^{a}_{k}) ds\ du, \label{eq: pIMU alpha noisy}\\
    %
    %
    %
    \breve{\pib}_{k+1} &\triangleq \Int{t_k}{t_{k+1}} {}^{k}\breve{\rot}_{s} (\breve{\accel}_s - \bar{\bias}^{a}_{k}) ds,\label{eq: pIMU beta noisy}\\
    %
    %
    %
    %
    \breve{\pig}_{k+1} &\triangleq \Mint_{t_k}^{t_{k+1}}\frac{1}{2}\Omg(\breve{\angvel}_s - \bar{\bias}^{\o}_{k}){}^k\breve{\quat}_s ds. \label{eq: pIMU gamma noisy}
\end{align}

In practice, the above integration can be realized by zero-order-hold (ZOH) or higher order methods (Runge-Kutta methods). During the period from $t_k$ to $t_{k+1}$, we have a sequence of $m$ IMU samples $(\breve{\angvel}_{0},\ \breve{\accel}_0)$, $(\breve{\angvel}_{1},\ \breve{\accel}_1)$, $\dots$, $(\breve{\angvel}_{m-1},\ \breve{\accel}_{m-1})$. Thus, if ZOH is chosen, we have the following recursive rule to update the pre-integrated measurements:

\begin{subequations} \label{eq: pIMU zoh}
    \begin{align}[left = \empheqlbrace]
      &\breve{\pia}_{n+1} = \breve{\pia}_{n} + \breve{\pib}_{n}\D \tau_n + \frac{1}{2}{}^k\breve{\rot}_{n}(\breve{\accel}_n - \bar{\bias}^a_k)\D \tau_n^2, \label{eq: pIMU alpha zoh rule}
      \\
      &\breve{\pib}_{n+1} = \breve{\pib}_{n} + {}^k\breve{\rot}_{n}(\breve{\accel}_n - \bar{\bias}^a_k)\D \tau_n, \label{eq: pIMU beta zoh rule}
      \\
      &\breve{\pig}_{n+1} = \breve{\pig}_{n}
      \circ
      \calE\l(\D\tau_n(\breve{\angvel}_n - \bar{\bias}^\o_k)\r), \label{eq: pIMU gamma zoh rule}
      \\
      &\breve{\pia}_{0} = \zero,\ \breve{\pib}_{0} = \zero,\ \breve{\pig}_{0} = \begin{bmatrix}1 &\zero_{1\times3}\end{bmatrix}^\top,
      \\
      &\D \tau_n = \tau_{n+1} - \tau_{n},\ n = \{0, 1 \dots m-1\},
    \end{align}
\end{subequations}
where $\tau_n$ is the time stamp of the $n$-th sample, and $\tau_0 = t_k$, $\tau_m = t_{k+1}$.
Note that the above equations cannot be considered a usable observation model yet, since the observations are still functions of the chosen bias point, i.e. $\breve{\pia}_{k+1} = \breve{\pia}_{k+1}(\bar{\bias}_{k})$, $\breve{\pib}_{k+1} = \breve{\pib}_{k+1}(\bar{\bias}_{k})$, $\breve{\pig}_{k+1} = \breve{\pig}_{k+1}(\bar{\bias}_{k})$, $\bar{\bias}_{k} \triangleq (\bar{\bias}^{\o}_{k},\ \bar{\bias}^{a}_{k})$.
Let $\breve{\bias}_{k} = (\breve{\bias}^{\o}_{k},\ \breve{\bias}^{a}_{k})$ be a numerical value of the variable $\bar{\bias}_k$, we can further approximate the observation model in \eqref{eq: preint alpha ideal} \eqref{eq: preint beta ideal} \eqref{eq: preint gamma ideal} as follows:
\begin{align}
    &\breve{\pia}_{k+1}\Big|_{\bar{\bias}_{k} = \breve{\bias}_{k}}
    + \vbf{A}_{k+1}^{\o}\D \bias^{\o}_{k}
    + \vbf{A}_{k+1}^{a} \D \bias^{a}_{k}
    \nonumber
    \\
    &=
    \rot_k^{-1}(\pos_{k+1} - \pos_k - \vel_k\Dt_k + \frac{1}{2}{}\grav\Dt^2_k) - \d\pia_{k+1},
    \label{eq: preint alpha approx}
    \\
    &\breve{\pib}_{k+1}\Big|_{\bar{\bias}_{k} = \breve{\bias}_{k}}
    + \vbf{B}_{k+1}^{\o}\D \bias^{\o}_{k}
    + \vbf{B}_{k+1}^{a} \D \bias^{a}_{k}
    \nonumber
    \\
    &=
    \rot_k^{-1} (\vel_{k+1} - \vel_k + \grav\Dt_k) - \d\pib_{k+1},
    \label{eq: preint beta approx}
    \\
    &\breve{\pig}_{k+1}\Big|_{\bar{\bias}_{k}^{\o} = \breve{\bias}_{k}^{\o}}
    \circ
    \begin{bmatrix} 1\\ \frac{1}{2}\vbf{C}_{k+1}^{\o}\D\bias^{\o}_{k}\end{bmatrix}
    \simeq \breve{\pig}_{k+1}\Big|_{\bar{\bias}_{k}^{\o} = \bias_{k}^{\o}}
    \nonumber
    \\
    &=
    \quat_k^{-1} \circ \quat_{k+1}
    \circ
    \begin{bmatrix} 1\\ -\frac{1}{2}\d\pith_{k+1}\end{bmatrix},
    \label{eq: preint gamma approx}
\end{align}
where $\d\pia_{k+1}$, $\d\pib_{k+1}$, $\d\pith_{k+1}$ are the measurement errors, and $\vbf{A}, \vbf{B}, \vbf{C}$ are the Jacobians of the IMU preintegrations evaluated at the bias point $\breve{\bias}_k$, i.e.
{
\begin{align*}
\begin{matrix*}[l]
    \vbf{C}_{k+1}^{\o}
    \triangleq 
    \Jcb^{\breve{\pig}(\breve{\bias}^\o_k + \D\bias^\o_k)}_{\D\bias^\o_k},
    %
    &{}^k\breve{\rot}_{k+1} \triangleq \qtoR(\breve{\pig}_{k+1}),
    \\
    \vbf{A}_{k+1}^{\o}
    \triangleq
    \frac{\partial\breve{\pia}_{k+1}(\bar{\bias}_{k}^{\o})}
         {\partial \bar{\bias}^{\o}_{k}}
    \Big|_{\bar{\bias}_{k}^{\o} = \breve{\bias}_{k}^{\o}},
    &\vbf{A}_{k+1}^{a}
    \triangleq
    \frac{\partial\breve{\pia}_{k+1}(\bar{\bias}_{k}^{a})}
         {\partial \bar{\bias}^{a}_{k}}
    \Big|_{\bar{\bias}_{k}^{a} = \breve{\bias}_{k}^{a}},
    \\
    \vbf{B}_{k+1}^{\o}
    \triangleq
    \frac{\partial\breve{\pib}_{k+1}(\bar{\bias}_{k}^{\o})}
         {\partial \bar{\bias}^{\o}_{k}}
    \Big|_{\bar{\bias}_{k}^{\o} = \breve{\bias}_{k}^{\o}},
    &\vbf{B}_{k+1}^{a}
    \triangleq
    \frac{\partial\breve{\pib}_{k+1}(\bar{\bias}_{k}^{a})}
         {\partial \bar{\bias}^{a}_{k}}
    \Big|_{\bar{\bias}_{k}^{a} = \breve{\bias}_{k}^{a}},
    \\
    \D {\bias}_{\o_k}
    \triangleq \bias^{\o}_{k} - \breve{\bias}^{\o}_{k}
    ,\  %
    &\D {\bias}_{a_k}
    \triangleq
    \bias^{a}_{k} - \breve{\bias}^{a}_{k}.
\end{matrix*}
\end{align*}
}In practice $\breve{\bias}_{k}$ can be chosen as $\hat{\bias}_{k}$ at the beginning of the optimization process. During the optimization when the bias estimate is updated, the preintegrated terms can be recalculated by either redoing the integrations \eqref{eq: pIMU alpha noisy}, \eqref{eq: pIMU beta noisy}, \eqref{eq: pIMU gamma noisy} with new bias point, or by the first order approximation on the left hand side of the approximate models \eqref{eq: preint alpha approx}, \eqref{eq: preint beta approx}, \eqref{eq: preint gamma approx}, depending on how much the bias estimate changes.


To fully stipulate the observation model, the Jacobians $\vbf{A}$, $\vbf{B}$, $\vbf{C}$ must be stated as well. Based on \eqref{eq: pIMU zoh}, they can be also computed via an iterative scheme. First, let us find the formula for $\vbf{C}_{k+1}^\o$. We rewrite \eqref{eq: pIMU gamma zoh rule} as
\begin{align} \label{eq: rot chain}
    {}^k\breve{\rot}_{n+1} &= \prod_{j=0}^n \Exp\l((\breve{\angvel}_{j}-\breve{\bias}_k^\o)\D \tau_{j}\r) \nonumber\\
                           &= {}^k\breve{\rot}_{n} \Exp\l((\breve{\angvel}_{n}-\breve{\bias}_k^\o)\D \tau_{n}\r) \nonumber\\
                           &= {}^k\breve{\rot}_{n} {}^{n}\breve{\rot}_{n+1}.
\end{align}
Then, we apply the approximation of each of the terms ${}^k\breve{\rot}_{n+1}$, ${}^k\breve{\rot}_{n}$, ${}^{n}\breve{\rot}_{n+1}$ under some small changes in the bias:
\begin{align*}
    &{}^k\breve{\rot}_{n+1}\Exp(\vbf{C}_{n+1}^{\o}\D\bias^{\o}_k) \dots
    \\
    &\stackrel{\eqref{eq: first order approx of rodrigues formula}}{\simeq}{}^k\breve{\rot}_{n}\Exp(\vbf{C}_{n}^{\o}\D\bias^{\o}_k)
    {}^n\breve{\rot}_{n+1}\Exp\l(-\Hrj_n\D\tau_n \D\bias^{\o}_k\r)
    \\
    &\stackrel{\eqref{eq: ER=RE}}{=}{}^k\breve{\rot}_{n+1}\Exp\l({}^n\breve{\rot}_{n+1}^{-1}\vbf{C}_{n}^{\o}\D\bias^{\o}_k\r)\Exp\l(-\Hrj_n\D\tau_n \D\bias^{\o}_k\r)
    \\
    &\stackrel{\eqref{eq: first order approx two disturbance}}{\simeq} {}^k\breve{\rot}_{n+1}\Exp\l({}^n\breve{\rot}_{n+1}^{-1}\vbf{C}_{n}^{\o}\D\bias^{\o}_k - \Hrj_n\D\tau_n\D\bias^{\o}_k\r).
\end{align*}
Hence, we multiply ${}^k\breve{\rot}_{n+1}^{-1}$ to the left and apply $\Log(\cdot)$ to both sides, we have:
\begin{align*}
    \vbf{C}_{n+1}^{\o} \simeq {}^n\breve{\rot}_{n+1}^{-1}\vbf{C}_{n}^{\o} - \Hrj_n\D\tau_n,
\end{align*}
where in reference to \eqref{eq: first order approx of rodrigues formula}, $\Hrj_n$ and $\vbf{C}^0_\o$ are defined as:
\begin{align}
    \Hrj_n = \Hrj\l(\l(\breve{\angvel}_n - \breve{\bias}_k^\o\r)\D\tau_n\r),\ \vbf{C}_{0}^{\o} = \zero_{1\times3}.
\end{align}

From \eqref{eq: pIMU alpha zoh rule} it can be easily seen that
\begin{align*}
    &\vbf{A}_{n+1}^{\o}
    =
    \vbf{A}_{n}^{\o} + \vbf{B}_{n}^{\o}\D \tau_n
    -\frac{1}{2}
    {}^k\breve{\rot}_{n}\sksym{(\breve{\accel}_n - \breve{\bias}_k^a)}\vbf{C}_{n}^{\o}\D\tau_n^2,
    \\
    &\vbf{A}_{n+1}^{a}
    =
    \vbf{A}_{n}^{a} + \vbf{B}_{n}^{a}\D \tau_n
    -\frac{1}{2}
    {}^k\breve{\rot}_{n}\D\tau_n^2,
    \\
    &\vbf{A}_{0}^{\o} = 0,\ \vbf{A}_{0}^{a} = 0.
\end{align*}

Similarly, from \eqref{eq: pIMU beta zoh rule}, we get
\begin{align*}
    &\vbf{B}_{n+1}^{\o}
    =
    \vbf{B}_{n}^{\o}
    -{}^k\breve{\rot}_{n}\sksym{(\breve{\accel}_n - \breve{\bias}_k^a)}\vbf{C}_{n}^{\o}\D\tau_n,
    \\
    &\vbf{B}_{n+1}^{a}
    =
    \vbf{B}_{n}^{a}
    -{}^k\breve{\rot}_{n}\D\tau_n,
    \\
    &\vbf{B}_{0}^{\o} = 0,\ \vbf{B}_{0}^{a} = 0.
\end{align*}

\subsubsection{Measurement error covariance}
The covariances of the IMU preintegration measurements are calculated based on a propagation scheme similar to \cite{eckenhoff2019closed, qin2018vins}.
Specifically, recall that we have defined the measurement errors of the pIMU terms in \eqref{eq: preint alpha approx}, \eqref{eq: preint beta approx}, \eqref{eq: preint gamma approx} as $\tilde{\I}_t \triangleq (\d\pia_t,\ \d\pib_t,\ \d\pith_t,\ \d\bias^\o_t,\ \d\bias^a_t)$ where $\g_t = \breve{\g}_t \circ (1,\ \frac{1}{2}\d{\pith}_t)$, $\pia_t = \breve{\pia}_t + \d{\pia}_t$, $\pib_t = \breve{\pib}_t + \d{\pib}_t$, $\bias^{\o}_{t} = \breve{\bias}^{\o}_{k} + \d{\bias}^{\o}_{t}$, $\bias^{a}_{t} = \breve{\bias}^{a}_{k} + \d{\bias}^{a}_{t}$, the continuous time dynamics of the error can be defined as
\begin{align}
    \dot{\tilde{\I}}_t
    &=
    {\vbf{F}}_t \tilde{\I}_t
    +
    {\vbf{G}}_t\vbf{\eta}_{t},
\end{align}
where $\vbf{F}_t$, $\vbf{G}_t$, $\vbf{\eta}_{t}$ are defined as
\begin{align}
    {\vbf{F}}_t &\triangleq
    \begin{bmatrix}
        \zero  &\ident  &\zero    &\zero &\zero\\
        \zero  &\zero   &-{}^k\breve{\rot}_t\sksym{\breve{\accel}_t - \breve{\bias}^{a}_{k}} &\zero  &-{}^k\breve{\rot}_t\\
        \zero  &\zero   &-\sksym{\breve{\angvel}_t  - \breve{\bias}^{\o}_{k}}    &-\ident &\zero\\
        \zero  &\zero   &\zero    &\zero &\zero\\
        \zero  &\zero   &\zero    &\zero &\zero\\
    \end{bmatrix}, \nonumber\\
    {\vbf{G}}_t &\triangleq
    \begin{bmatrix}
        \zero    &\zero     &\zero    &\zero\\
        \zero    &-{}^k\breve{\rot}_t &\zero  &\zero\\
        \zero    &-\ident   &\zero    &\zero\\
        \zero    &\zero     &\ident   &\zero\\
        \zero    &\zero     &\zero    &\ident\\
    \end{bmatrix},\ 
    \vbf{\eta}_{t} \triangleq
    \begin{bmatrix}{\bm\eta}^{\o}_{t},\\ {\bm\eta}^{a}_{t},\\ {\bm\eta}^{b\o}_{t},\\ {\bm\eta}^{ba}_{t}\end{bmatrix}.
\end{align}

Hence, if we assume that the angular velocity and acceleration measurements are constant between two consecutive IMU sample times $\tau_n$ and $\tau_{n+1}$, then ${\vbf{F}}_t$ and ${\vbf{G}}_t$ are constant in $[\tau_n,\ \tau_{n+1}]$. Thus the discrete time dynamics of the measurement error can be calculated by
\begin{align}
    \tilde{\I}_{n+1} &= \exp(\vbf{F}_n\D\tau_n)\tilde{\I}_{n} \dots \nonumber\\
                     &\qquad\qquad\qquad+ \Int{0}{\D\tau_{n}}\exp({\vbf{F}}_n s){\vbf{G}}_n \vbf{\eta}_{s + \tau_n} ds, 
\end{align}
and the covariance of $\tilde{\I}_t$ can be propagated from time $t_k$ to $t_{k+1}$ via the following iterative steps
\begin{align}
    &{\vbf{P}}_n = \zero,\ \tau_n = t_k,\\
    &{\vbf{P}}_{n+1} = \exp({\vbf{F}}_n\D \tau_n){\vbf{P}}_{n}\exp({\vbf{F}}_n\D \tau_n)^\top + {\vbf{Q}}_n,\\
    &{\vbf{Q}}_n = \Int{0}{\D \tau_n}\exp({\vbf{F}}_n s){\vbf{G}}_n {\vbf{Q}} {\vbf{G}}_n^\top \exp({\vbf{F}}_n s)^\top ds,
\end{align}
where ${\vbf{Q}} \triangleq \mathrm{diag}\{\s_{\vbf{\eta}_\o}^2, \s_{\vbf{\eta}_a}^2, \s_{\vbf{\eta}_{b\o}}^2, \s_{\vbf{\eta}_{ba}}^2\}$, and recall that $\exp(\vbf{F}_n\D \tau_n)$ is the matrix exponential of $\vbf{F}_n\D \tau_n$. This matrix exponential and $\vbf{G}_n$, $\vbf{Q}_n$ can be computed by several techniques such as first-order approximation $\exp(\vbf{F}_n\D s) \simeq \ident + s\vbf{F}_n$ in \cite{qin2017vins} or closed-form in \cite{eckenhoff2019closed}. In this work we use the first-order approximation form.



\subsubsection{Residual}
The pIMU residual $\resi_{\I}(\hat{\X}_k, \hat{\X}_{k+1}, \I_k)$ is therefore defined as
\begin{align}
    &\resi_{\I}(\hat{\X}_k, \hat{\X}_{k+1}, \I_k)
    \triangleq
    (
        \resi_{\pig},\
        \resi_{\pia},\ 
        \resi_{\pib},\ 
        \resi_{b\o},\ 
        \resi_{ba}
    ),
    \\
    &\resi_{\pig}
    \triangleq
    2\vbf{vec}
    \Bigg(
        %
        \l(\breve{\pig}_k \circ \calE(\vbf{C}^{k}_{\o}\D\hat{\bias}^{\o}_{k})\r)^{-1}
        \circ
        \D\hat{\quat}_{k}
    \Bigg),\\
    &\resi_{\pia}
    \triangleq
    \hat{\rot}_k^{-1}(\hat{\pos}_{k+1} - \hat{\pos}_k - \hat{\vel}_k\Dt_k + \frac{1}{2}{}\grav\Dt^2_k)+\dots
    \nonumber\\
    &\qquad\qquad
    -\vbf{A}_{k+1}^{\o} \D \hat{\bias}^{\o}_{k}
    -\vbf{A}_{k+1}^{a} \D \hat{\bias}^{a}_{k}
    -\breve{\pia}_{k+1},
    \\
    &\resi_{\pib}
    \triangleq
    \hat{\rot}_k^{-1}(\hat{\vel}_{k+1} - \hat{\vel}_k + \grav\Dt_k)+\dots
    \nonumber\\
    &\qquad\qquad - \vbf{B}_{k+1}^{\o} \D \hat{\bias}^{\o}_{k}
    -\vbf{B}_{k+1}^{a} \D \hat{\bias}^{a}_{k} - \breve{\pib}_k,
    \\
    &\resi_{b\o}
    \triangleq
    \hat{\bias}^{\o}_{k+1} - \hat{\bias}^{\o}_{k},
    \ 
    \resi_{ba}
    \triangleq
    \hat{\bias}^{a}_{k+1} - \hat{\bias}^{a}_{k}, 
\end{align}
where $\D \hat{\bias}^{\o}_{k} \triangleq \hat{\bias}^{\o}_{k} - \breve{\bias}^{\o}_{k}$, $\D \hat{\bias}^{a}_{k} \triangleq \hat{\bias}^{a}_{k} - \breve{\bias}^{a}_{k}$, $\D\hat{\quat}_{k} \triangleq \hat{\quat}_k^{-1} \circ \hat{\quat}_{k+1}$.

\subsubsection{Jacobian}
Following the same manipulations in the case of the OSL factor, we can easily find the Jacobians of the preintegrated quaternion residual $\resi_{\pig}$ over the state estimates:
\begin{alignat}{3}
    &\Jcb^{\resi_\pig}_{\hat{\quat}_k}
    &&=
    -&&\brc_3\big( \qmatR{\tilde{\pig}_k} \qmatL{\hat{\quat}_{k+1}^{-1} \circ \hat{\quat}_{k}} \big),
    \\
    &\Jcb^{\resi_\pig}_{\hat{\quat}_{k+1}}
    &&=
    &&\brc_3\l( \qmatL{ (\tilde{\pig}_k)^{-1} \circ \hat{\quat}_{k}^{-1} \circ \hat{\quat}_{k+1}} \r),
    \\
    &\Jcb^{\resi_\pig}_{\hat{\bias}_{{k}^\o}}
    &&=
    -&&\brc_3\l( \qmatR{ (\breve{\pig}_k)^{-1} \circ \hat{\quat}_{k}^{-1} \circ \hat{\quat}_{k+1}} \r)\vbf{C}^{k}_{\o},
\end{alignat}
where $\tilde{\pig}_k \triangleq \breve{\pig}_k \circ \calE\l( \vbf{C}^{k}_{\o}\D\hat{\bias}^{\o}_{k}\r)$.

The Jacobians of the preintegrated quaternion residual $\resi_{\pia}$ over the state estimates are
\begin{alignat}{2}
    &\Jcb^{\resi_\pia}_{\hat{\quat}_k}
    =
    &&\Bigg\lfloor\hat{\rot}_k^\top\Big(\hat{\pos}_{k+1} - \hat{\pos}_k - \hat{\vel}_k\Dt_k + \frac{\Dt^2_k}{2}{}\grav \Big)\Bigg\rfloor_{\times},
    \\
    &\Jcb^{\resi_\pia}_{\hat{\pos}_{k}}
    =
    -&&\hat{\rot}_k^\top,
    \mkern33mu
    \Jcb^{\resi_\pia}_{\hat{\pos}_{k+1}}
    =
    \hat{\rot}_{k}^\top,
    \quad
    \Jcb^{\resi_\pia}_{\hat{\vel}_{k}}
    =
    -\hat{\rot}_{k}^\top\Dt_k,
    \\
    &\Jcb^{\resi_\pia}_{\hat{\bias}^{\o}_{k}}
    =
    -&&\vbf{A}^{k+1}_\o,
    \quad
    \Jcb^{\resi_\pia}_{\hat{\bias}^{a}_{k}}
    =
    -\vbf{A}^{k+1}_a.
\end{alignat}

The Jacobians of the preintegrated quaternion residual $\resi_{\pib}$ over the state estimates are found as
\begin{alignat}{2}
    &\Jcb^{\resi_\pib}_{\hat{\quat}_k}
    =
    \sksym{\hat{\rot}_k^\top\Big(\hat{\vel}_{k+1} - \hat{\vel}_k + \grav\Dt_k\Big)},
    &&\\
    &\Jcb^{\resi_\pib}_{\hat{\vel}_{k}}
    =
    -\hat{\rot}_k^\top,
    \qquad\quad
    \Jcb^{\resi_\pib}_{\hat{\vel}_{k+1}}
    =
    \hat{\rot}_{k}^\top,
    \\
    &\Jcb^{\resi_\pib}_{\hat{\bias}^{\o}_{k}}
    =
    -\vbf{B}^{k}_\o,
    \qquad\quad
    \Jcb^{\resi_\pib}_{\hat{\bias}^{a}_{k}}
    =
    -\vbf{B}^{k}_a.
\end{alignat}

Finally, for $\resi_{b\o}$, $\resi_{ba}$, it is straightforward that
\begin{align}
        \Jcb^{\resi_{b\o}}_{\hat{\bias}^{\o}_{k}}
        =
        -\ident_{3\times3},\qquad\quad
        &\Jcb^{\resi_{b\o}}_{\hat{\bias}_{k+1}^\o}
        =
        \ident_{3\times3},
        \\
        \Jcb^{\resi_{ba}}_{\hat{\bias}^{a}_{k}}
        =
        -\ident_{3\times3},\qquad\quad
        &\Jcb^{\resi_{ba}}_{\hat{\bias}_{k+1}^a}
        =
        \ident_{3\times3}.
\end{align}

\begin{rem}
The above calculations can be found in 
\end{rem}

\subsection{UWB factors} \label{sec: range factors}

At each time instance $t^i \in (t_{k},\ t_{k+1}]$ we have the following measurements and priori from a UAV range:
\begin{align}
    \U^i &= \l(\breve{d}^i,\ \vbf{x}^i,\ \vbf{y}^i,\ \dt^i, \Dt_k\r),\ i \in N_\U^k
\end{align}
where $\breve{d}^i$ is the distance measurement, $\Dt_k \triangleq t_{k+1} - t_{k}$, $\dt^i \triangleq t^i - t_k$, $\vbf{x}^i$ is the position of the UWB anchor \wrt the world frame, and $\vbf{y}^i$ is the coordinate of the UWB ranging node in the body frame.
The UWB factor of this measurement is constructed via the following steps.

\subsubsection{Observation model}
If we assume that the velocity and orientation of the robot change at a constant rate from time $t_{k}$ to $t_{k+1}$, then at time $t_{k} + \dt^i$, we have the following relative position of a UAV ranging node $\vbf{y}^i$ from an anchor $\vbf{x}^i$ as
\begin{align}
    \vbf{n}^i   &= \vbf{n}(\pos_{k+1},\ \quat_{k},\ \quat_{k+1},\ \vel_{k},\ \vel_{k+1},\ \dt^i,\ \Dt_k) \nonumber \\
                &\triangleq \pos_{k+1} + \rot_{k}\Exp\l(\frac{\dt^i}{\Dt_k}\Log(\rot_{k}^{-1}\rot_{k+1})\r) \vbf{y}^i \nonumber \\
                &\qquad - \int_{\dt^i}^{\Dt_k} \l(\vel_{k} + \frac{\tau}{\Dt_{k+1}}(\vel_{k+1}-\vel_{k})\r)d\tau - \vbf{x}^i \nonumber \\
                &\triangleq \pos_{k+1} + \rot_{k}\Exp\l(s^i\Log(\rot_{k}^{-1}\rot_{k+1})\r) \vbf{y}^i \nonumber\\
                &\qquad\qquad\qquad\qquad\qquad + a^i\vel_{k+1} + b^i{\vel}_{k} - \vbf{x}^i, \label{eq: ant-anc displacement}
\end{align}
where
\begin{align}
    s^i = \frac{\dt_i}{\Dt_k},\ &a_k^i = -\frac{\Dt_k^2 - \dt_i^2}{2\D t_k},\ b_k^i = -\frac{(\Dt_k - \dt_i)^2}{2\Dt_k}.
\end{align}

Hence, we consider the distance measurement $\breve{d}^i$ at time $t_k+\dt^i$ as the norm of the vector $\vbf{n}^i$, corrupted by a zero-mean Gaussian noise $\vbf{\eta}_{\U^i} \sim \mathcal{N}(0, \s_{\U}^2)$:
\begin{align}
    \breve{d}^i &= \norm{\vbf{n}(\pos_{k+1},\ \quat_{k},\ \quat_{k+1},\ \vel_{k},\ \vel_{k+1})} + \vbf{\eta}_{\U^i} \label{eq: distance model}
\end{align}

\subsubsection{Observation error covariance}

Based on \eqref{eq: distance model}, we can straightforwardly identify the covariance of the measurement error as
\begin{equation}
    \vbf{P}_{\U^i} \equiv \s_{\U}^2.
\end{equation}

\subsubsection{Residual}
The residual of a range measurement $\U_k$ is the difference between the distance estimate $\hat{d}_k$ and the distance measurement $\breve{d}_k$:
\begin{equation}\label{eq: uwb residual}
    \resi_{\U}(\hat{\X}_{k}, \hat{\X}_{k+1}, \U^i) \triangleq \resi_{\U^i} = \| \hat{\vbf{n}}^i\| - \breve{d}^i,
\end{equation}
where $\hat{\vbf{n}}^i = \vbf{n}(\hat{\pos}_{k+1},\ \hat{\quat}_{k},\ \hat{\quat}_{k+1},\ \hat{\vel}_{k},\ \hat{\vel}_{k+1})$.

\subsubsection{Jacobian}
From \eqref{eq: uwb residual} and \eqref{eq: ant-anc displacement} the Jacobians of the UWB residual over the position and velocity states can be easily calculated as follows
\begin{equation}
    \Jcb^{\resi_{\U^i}}_{\hat{\pos}_{k+1}} 
    =  (\hat{\vbf{n}}^i)^\top,\
    \Jcb^{\resi_{\U^i}}_{\hat{\vel}_{k+1}} 
    = a^i(\hat{\vbf{n}}^i)^\top,\
    \Jcb^{\resi_{\U^i}}_{\hat{\vel}_{k}} 
    = b^i(\hat{\vbf{n}}^i)^\top.
\end{equation}

The Jacobians of the orientation states can be found by using the chain rule. First let us consider the variable $\hat{\quat}_{k+1}$:
\begin{equation} \label{eq: jacobian Ru over Rk+1}
    \Jcb^{\resi_{\U^i}}_{\hat{\quat}_{k+1}}
    =
    \Jcb^{\resi_{\U^i}}_{\bar{\rot}^i\vbf{y}}
    \Jcb^{\bar{\rot}^i\vbf{y}}_{\bar{\rot}^i}
    \Jcb^{\bar{\rot}^i}_{s^i\bar{\eulvec}_k}
    \Jcb^{s^i\bar{\eulvec}_k}_{\hat{\quat}_{k+1}},
\end{equation}
where $\bar{\rot}^i \triangleq \Exp(s^i\bar{\eulvec}_k)$ and $\bar{\eulvec}_k \triangleq \Log(\hat{\rot}_{k}^{-1}\hat{\rot}_{k+1})$, and the intermediate Jacobians in the above chain are
\begin{alignat}{2}
    &\Jcb^{\resi_{\U^i}}_{\bar{\rot}^i\vbf{y}}
    = (\hat{\vbf{n}}^i)^\top\hat{\rot}_{k},
    \qquad
    &&\Jcb^{\bar{\rot}^i\vbf{y}}_{\bar{\rot}^i}
    = -\Exp(s^i\bar{\eulvec}_k)\sksym{\vbf{y}},
    \\
    &\Jcb^{\bar{\rot}^i}_{s^i\bar{\eulvec}_k}
    = \Hrj(s^i\bar{\eulvec}_k),
    &&\Jcb^{s^i\bar{\eulvec}_k}_{\hat{\quat}_{k+1}}
    = s^i \Hrj^{-1}(\bar{\eulvec}_k)
\end{alignat}

To find the Jacobian of ${\resi}_{\U^i}$ with respect to $\hat{\quat}_{k}$, some manipulations are needed. First we notice that 
\begin{align*}
    \bar{\rot}^i &= \hat{\rot}_{k+1} \hat{\rot}_{k+1}^{-1}\hat{\rot}_{k}\Exp(s^i\bar{\eulvec}_k)\\
                 &= \hat{\rot}_{k+1} \Exp(\Log(\hat{\rot}_{k+1}^{-1}\hat{\rot}_{k}))\Exp(s^i\bar{\eulvec}_k)\\
                 &= \hat{\rot}_{k+1} \Exp((s^i-1)\bar{\eulvec}_k) \triangleq \hat{\rot}_{k+1} \Exp(\bar{s}^i\bar{\eulvec}_k).
\end{align*}
And thus we can obtain a similar Jacobian compared to \eqref{eq: jacobian Ru over Rk+1}, except for $\bar{s}^i$ in place of $s^i$:
\begin{equation} \label{eq: jacobian Ru over Rk}
    \Jcb^{\resi_{\U^i}}_{\hat{\quat}_{k}}
    =
    \Jcb^{\resi_{\U^i}}_{\bar{\rot}^i\vbf{y}}
    \Jcb^{\bar{\rot}^i\vbf{y}}_{\bar{\rot}^i}
    \Jcb^{\bar{\rot}^i}_{\bar{s}^i\bar{\eulvec}_k}
    \Jcb^{\bar{s}^i\bar{\eulvec}_k}_{\hat{\quat}_{k}},
\end{equation}
where the intermediate Jacobians are
\begin{alignat}{2}
    &\Jcb^{\resi_{\U^i}}_{\bar{\rot}^i\vbf{y}}
    = (\hat{\vbf{n}}^i)^\top\hat{\rot}_{k},
    \qquad
    &&\Jcb^{\bar{\rot}^i\vbf{y}}_{\bar{\rot}^i}
    = -\Exp(\bar{s}^i\bar{\eulvec}_k)\sksym{\vbf{y}}
    \\
    &\Jcb^{\bar{\rot}^i}_{\bar{s}^i\bar{\eulvec}_k}
    = \Hrj(\bar{s}^i\bar{\eulvec}_k),
    &&\Jcb^{\bar{s}^i\bar{\eulvec}_k}_{\hat{\quat}_{k}}
    = \bar{s}^i \Hrj^{-1}(\bar{\eulvec}_k)
\end{alignat}


\section{Experiments} \label{sec: experiment}

In this section, we report results of the sensor fusion method on three types of datasets: public datasets, high-fidelity graphical-physical simulation, and field-collected datasets. Some implementation issues will also be discussed. Video illustrating the sensor fusion results on these datasets can be viewed at \url{https://youtu.be/2um-n_Wc9-k}.

\subsection{Public datasets} \label{sec: pub dataset}

\subsubsection{Preparation} \label{sec: pub dataset preperation}
Before working with physical systems, we employ the EuRoC datasets\cite{burri2016euroc} to verify the efficacy of our sensor fusion method. We only focus on the V1 and V2 datasets, as these experiments provide both position and orientation ground truth data from VICON, which are necessary to demonstrate the capability of our method to estimate not only the position but also the orientation.

Since the datasets do not contain UWB measurements, vicon data are used to generate UWB measurements, with realistic features taken into account. Specifically we simulate a UWB network of four UWB anchor nodes and two UAV nodes. In addition, each UAV node is assumed to have two antennae A and B, thus there are effectively four UAV ranging nodes in total. We assign the ID numbers 100, 101, 102, 103 to the anchors, and their respective coordinates in the world frame are chosen as (3, 3, 3), (3, -3, 0.5), (-3, -3, 3), (-3, 3, 0.5). Similarly, we assign the ID numbers 200.A, 200.B, 201.A, 201.B to the UAV ranging nodes, and their coordinates in the body frame are chosen as (0.25, -0.25, 0), (0.25, 0.25, 0), (-0.25, 0.25, 0), (-0.25, -0.25, 0).

Each UAV ranging node is given a pre-determined ranging cycle. Specifically, the ranging sequence of node 200 is set as:
\[\text{200.A} \to \text{100},\ \text{200.B} \to \text{100},\ \text{200.A} \to \text{101},\ \text{200.B} \to \text{101},\]
\[\text{200.A} \to \text{102},\ \text{200.B} \to \text{102},\ \text{200.A} \to \text{103},\ \text{200.B} \to \text{103}.\]
Concurrently, the ranging cycle of 201 is similar, but the ID of the anchor node is cyclically shifted by two units, i.e.:
\[\text{201.A} \to \text{102},\ \text{201.B} \to \text{102},\ \text{201.A} \to \text{103},\ \text{201.B} \to \text{103},\]
\[\text{201.A} \to \text{100},\ \text{201.B} \to \text{100},\ \text{201.A} \to \text{101},\ \text{201.B} \to \text{101}.\]

Hence, at each step, two UWB measurements can be obtained simultaneously, over a period of 6 steps. The steps are set at 25 ms apart, and we will effectively have an 80 Hz update rate for UWB measurements. Finally a zero-mean Gaussian noise with the standard deviation of 0.05m is added to the measurement, based on the characteristics of the physical sensor.

The software packages are implemented under ROS framework, and the optimization method employs the ceres solver. For each dataset, we run the VINS-Fusion\footnote{\url{https://github.com/HKUST-Aerial-Robotics/VINS-Fusion}} algorithm and
ORB-SLAM3\footnote{\url{https://github.com/UZ-SLAMLab/ORB_SLAM3}} \cite{campos2020orb} to estimate the aerial robot's pose based on data from the stereo camera and IMU. The VINS-Fusion estimate is then treated as an OSL data and fused with UWB and IMU using the methods proposed in previous sections. 

\subsubsection{Results and evaluation}

After running the algorithms on the datasets, we can obtain the OSL-based trajectory ${}^{\fL}\textbf{X}_\mathrm{OSL} = \{({}^{\fL}\breve{\pos}_k, {}^{\fL}\breve{\quat}_k)\}_{k=0}^K$ from the OSL system, and ${}^{\fW}\textbf{X}_\mathrm{VIRAL} = \{({}^{\fW}\hat{\pos}_k, {}^{\fW}\hat{\quat}_k)\}_{k=0}^K$ from the VIRAL method.
Since each OSL method provides a trajectory estimate ${}^{\fL}\textbf{X}_\mathrm{OSL}$ \wrt to a local frame $\{\fL\}$ that coincides with the initial pose $({}^{\fW}\pos_0, {}^{\fW}\quat_0)$ of the UAV, we need to transform ${}^{\fL}\textbf{X}_\mathrm{OSL}$ to ${}^{\fW}\textbf{X}_\mathrm{OSL} = \{({}^{\fW}\breve{\pos}_k, {}^{\fW}\breve{\quat}_k)\}_{k=0}^K$ before we can compare it with ${}^{\fW}\textbf{X}_\mathrm{VIRAL}$. To this end, we perform:
\begin{align}[left = \empheqlbrace]
      {}^{\fW}\breve{\pos}_k  &= \qtoR({}^{\fW}\quat_0) {}^{\fL}\breve{\pos}_k + {}^{\fL}\pos_0,\\
      {}^{\fW}\breve{\quat}_k &= {}^{\fW}\quat_0\circ {}^{\fL}\breve{\quat}_k,
\end{align}
where $({}^{\fW}\pos_0, {}^{\fW}\quat_0)$ is the initial pose of the UAV according to ground truth. Hence we calculate the Root Mean Square Error for the position and rotation estimates as follows
\begin{align}
    \text{RMSE}_\text{pos} &= \sqrt{\frac{1}{N+1}\sum_{k=0}^K\norm{\pos_k - \bar{\pos}_k}^2},\\
    \text{RMSE}_\text{rot} &= \sqrt{\frac{1}{N+1}\sum_{k=0}^K\Log(\bar{\rot}^{-1}_k\rot_k)},
\end{align}
where ${\pos}_k$ and ${\rot}_k$ are the position and orientaion from groundtruth, and $\bar{\pos}_k$ and $\bar{\rot}_k$ are the position and orientation estimate from either OSL or VIRAL algorithm. The result is shown in Tab. \ref{tab: euroc ATE}.

\begin{figure*}
	\centering
	\begin{subfigure}[h]{0.3125\linewidth}
        \centering
        \includegraphics[width=\linewidth]{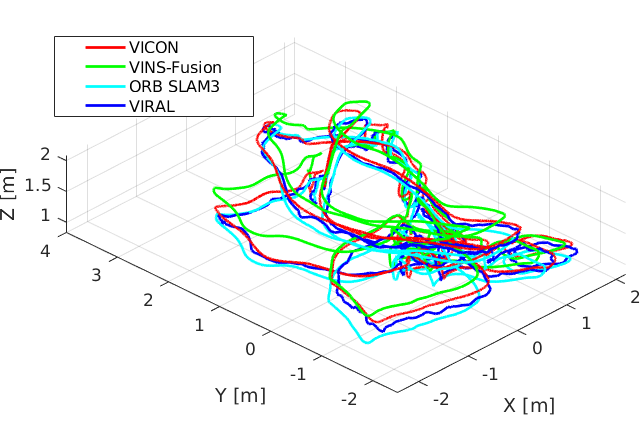}
		\caption{V1\_01 dataset.}
		\label{fig: euroc xyz v101}
	\end{subfigure}
    \begin{subfigure}[h]{0.3125\linewidth}
        \centering
		\includegraphics[width=\linewidth]{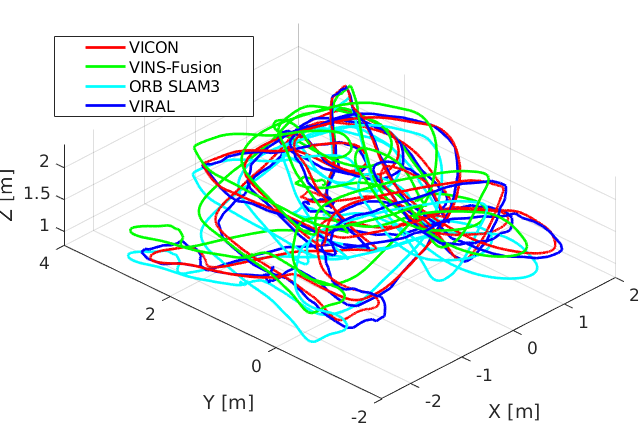}
		\caption{V1\_02 dataset.}
		\label{fig: euroc xyz v102}
	\end{subfigure}
	\begin{subfigure}[h]{0.3125\linewidth}
        \centering
        \includegraphics[width=\linewidth]{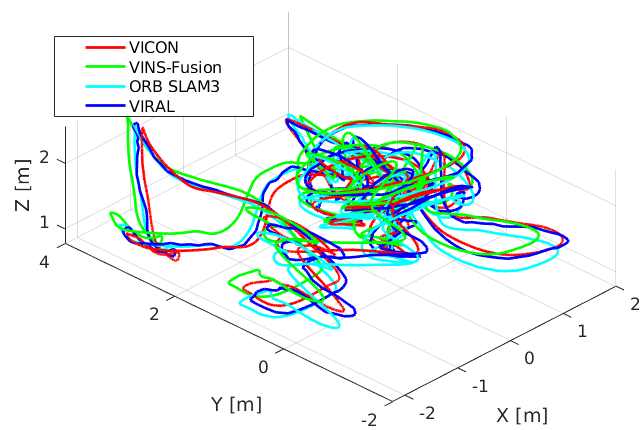}
        \caption{V1\_03 dataset.}
		\label{fig: euroc xyz v103}
	\end{subfigure}
	
	\begin{subfigure}[h]{0.3125\linewidth}
        \centering
        \includegraphics[width=\linewidth]{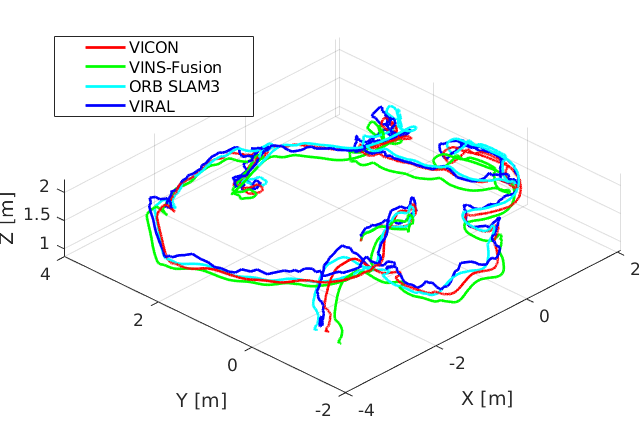}
		\caption{V2\_01 dataset.}
		\label{fig: euroc xyz v201}
	\end{subfigure}
    \begin{subfigure}[h]{0.3125\linewidth}
        \centering
		\includegraphics[width=\linewidth]{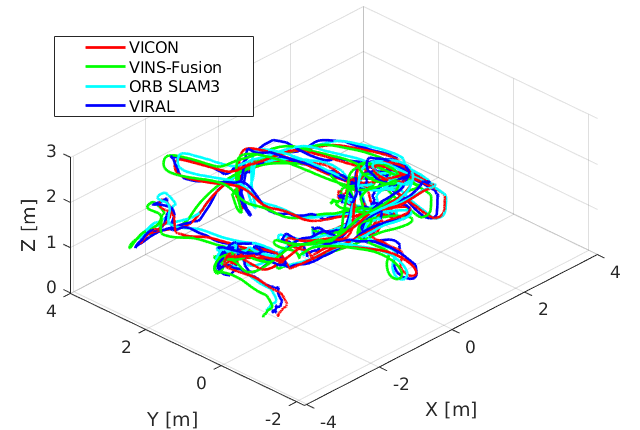}
		\caption{V2\_02 dataset.}
		\label{fig: euroc xyz v202}
	\end{subfigure}
	\begin{subfigure}[h]{0.3125\linewidth}
        \centering
        \includegraphics[width=\linewidth]{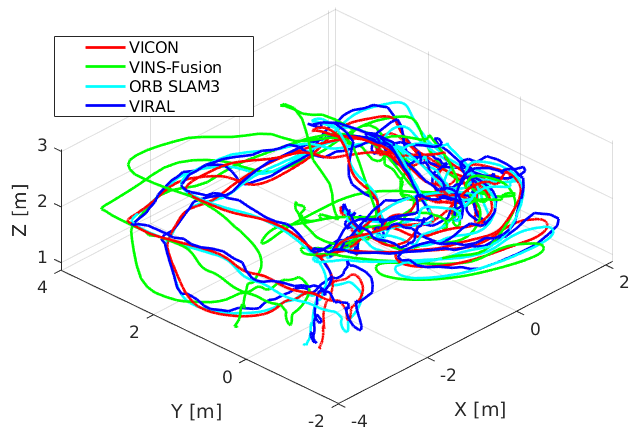}
        \caption{V2\_03 dataset.}
		\label{fig: euroc xyz v203}
	\end{subfigure}

	\caption{Trajectories of the UAV and the estimates by multiple methods with the EuRoC datasets.}
	\label{fig: euroc xyz}
\end{figure*}

\begin{figure}[t]
    \centering
    \includegraphics[width=\linewidth]{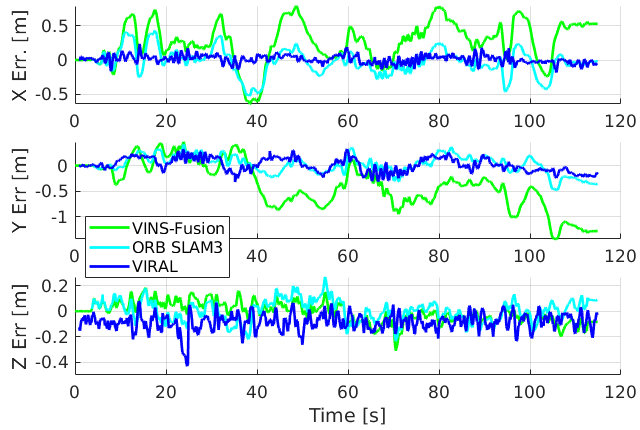}
    \caption{Position estimation error of OSL and VIRAL methods on the V1\_03 dataset.}
    \label{fig: euroc err xyz t}
\end{figure}

\begin{figure}[t]
    \centering
    \includegraphics[width=0.98\linewidth]{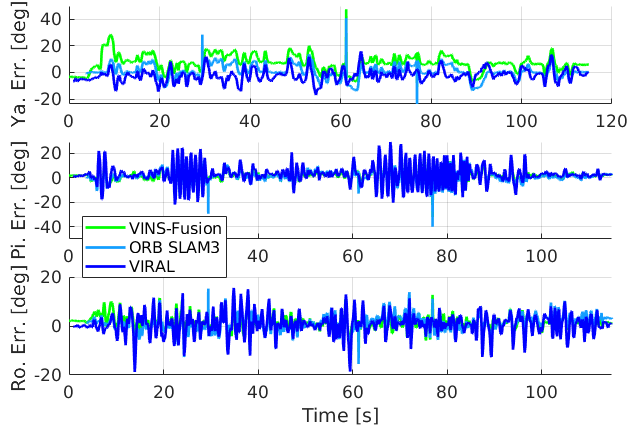}
    \caption{Orientation estimation error of OSL and VIRAL methods from the V1\_03 dataset.}
    \label{fig: euroc err ypr t}
\end{figure}

\begin{table}[t]
    \setlength{\tabcolsep}{3.75pt}
    \centering
    \renewcommand{\arraystretch}{1.1}
	\caption{Translational and rotational RMSE of VINS-Fusion, ORB-SLAM3 and VIRAL estimates on the EuRoC datasets.}
	\label{tab: euroc ATE}
    \begin{tabu} to \textwidth {c||c|c|c||c|c|c}
    \hline\hline
                        &\mc{3}{c||}{$\text{RMSE}_\text{pos}$ [m]}
                        &\mc{3}{c}{$\text{RMSE}_\text{rot}$ [deg]}\\\cline{2-7}
    \mr{-2}{*}{Dataset} &VINS      &ORB3     &VIRAL         &VINS          &ORB3         &VIRAL\\\hline
    V1\_01              &0.2508    &0.2230   &\bf{0.1442}   &\bf{0.7935}   &2.7362       &2.3019\\
    V1\_02              &0.3215    &0.3628   &\bf{0.1509}   &\bf{1.7794}   &2.8454       &2.7633\\
    V1\_03              &0.2460    &0.1937   &\bf{0.1498}   &5.9205        &3.3969       &\bf{1.9597}\\
    V2\_01              &0.1948    &0.1891   &\bf{0.1482}   &6.9295        &6.0469       &\bf{5.6066}\\
    V2\_02              &0.3588    &0.2569   &\bf{0.1685}   &8.8905        &\bf{8.4701}  &9.5030\\
    V2\_03              &0.7027    &0.2630   &\bf{0.1663}   &11.9267       &\bf{8.9584}  &10.2057\\\hline\hline
    \end{tabu}
\end{table}

From Tab. \ref{tab: euroc ATE}, we can clearly see that the VIRAl scheme achieves better positional RMSE compared to both VINS and ORB-SLAM3. This is the desired and expected outcome, since the VINS-Fusion and ORB-SLAM3 algorithms, being OSL methods, will accumulate error over time, while the VIRAL method employs ranging measurements that can constrain the error. This can be seen more clearly in Fig. \ref{fig: euroc xyz} and Fig. \ref{fig: euroc err xyz t}, where the trajectory estimate by VINS-Fusion clearly diverges from the ground truth towards the end of the trajectory, while the VIRAL estimate still track the ground truth.

In terms of rotational RMSE, we can see from Tab. \ref{tab: euroc ATE} that our proposed scheme can work well in the same number of datasets as VIO methods. The same explanation for the positional RSME can be applied here, as the body-offset ranging scheme can ensure that estimation error will not drift, while the OSL method may suffer from some bias error, as can be seen from Fig. \ref{fig: euroc err ypr t}.

\begin{figure*}
	\centering
	\begin{subfigure}[h]{0.3125\linewidth}
        \centering
        \includegraphics[width=\linewidth]{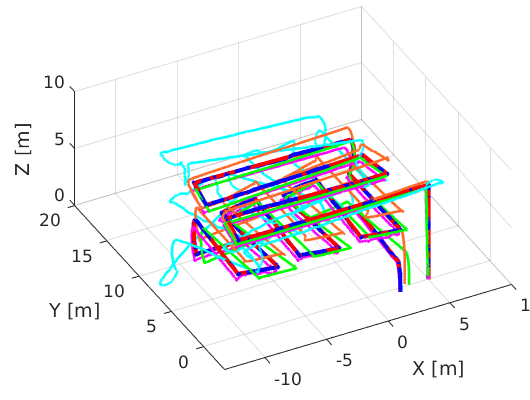}
		\caption{Dataset \#01.}
		\label{fig: airsim xyz interwoven}
	\end{subfigure}
    \begin{subfigure}[h]{0.3125\linewidth}
        \centering
		\includegraphics[width=\linewidth]{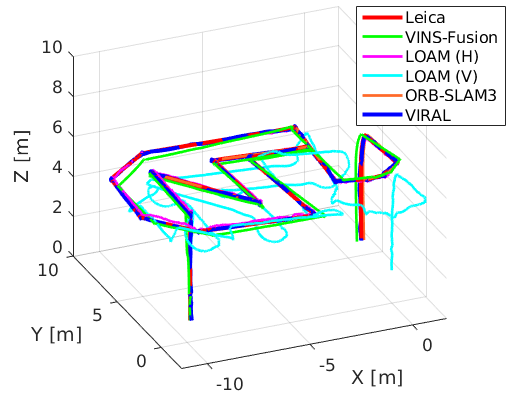}
		\caption{Dataset \#02.}
		\label{fig: airsim xyz ntu}
	\end{subfigure}
	\begin{subfigure}[h]{0.3125\linewidth}
        \centering
        \includegraphics[width=\linewidth]{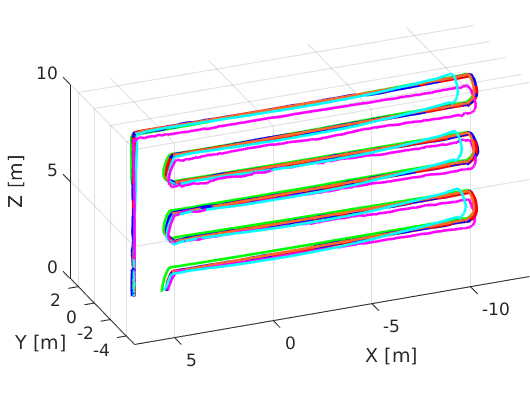}
        \caption{Dataset \#03.}
		\label{fig: airsim xyz xsweep}
	\end{subfigure}

	\caption{Trajectories of the UAV and the estimates from multiple methods in the datasets generated from AirSim.}  \label{fig: airsim xyz}
\end{figure*}

\begin{table*}[h]
    \setlength{\tabcolsep}{3.75pt}
    \centering
    \renewcommand{\arraystretch}{1.1}
	\caption{Translational and rotational RMSE of VINS-Fusion, LOAM and VIRAL estimates on our AirSim datasets.}
	\label{tab: airsim ATE}
    \begin{tabu} to \textwidth {c||c|c|c|c|c||c|c|c|c|c||c|c|c}
    \hline\hline
                        &\mc{5}{c||}{$\text{RMSE}_\text{pos}$ [m]}
                        &\mc{5}{c||}{$\text{RMSE}_\text{rot}$ [deg]}
                        &\mc{3}{c}{$\text{RMSE}_\text{vel}$ [ms/s]}\\\cline{2-14}
    \mr{-2}{*}{Dataset} &VINS &ORB3 &LOAM (H) &LOAM (V) &VIRAL
                        &VINS &ORB3 &LOAM (H) &LOAM (V) &VIRAL
                        &VINS &ORB3 &VIRAL\\\hline
    01 &0.8616 &1.1260 &0.5561 &4.5228  &\bf{0.0736}   
       &5.7186 &3.5593 &1.2890 &17.7741 &\bf{1.2291}   
       &0.1071 &0.2633 &\bf{0.0699}\\                  
    02 &0.4735 &0.1102 &0.1283 &2.1169 &\bf{0.0656}    
       &4.1718 &\bf{0.1561} &1.0533 &2.3456 &2.7413    
       &\bf{0.0824} &0.2613 &0.0830\\                  
    03 &0.8392 &0.6343 &0.3427 &0.3902 &\bf{0.1222}    
       &1.6726 &\bf{0.2957} &1.6611 &1.9817 &1.4319    
       &0.2352 &0.4385 &\bf{0.1438}\\\hline\hline      
    \end{tabu}
\end{table*}

\begin{figure}[h]
    \centering
    \includegraphics[width=\linewidth]{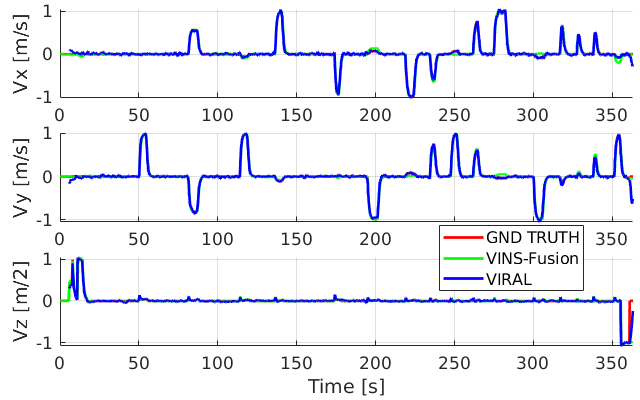}
	\caption{Velocity estimate from VINS-Fusion and VIRAL compared with groundtruth from the AirSim test \#02.}
	\label{fig: airsim vel}
\end{figure}

\subsection{High fidelity visual and physical simulation}

Though the EuRoC dataset has demonstrated the effectiveness of our localization method, it does not match well with our experiment setup in later part. Specifically, the movement is limited within a small area, and the dataset does not contain Lidar measurements. Moreover, the dataset does not contain velocity groundtruth to compare with our estimate. Hence, to verify the method in a more general manner, we employ the AirSim software package \cite{shah2018airsim} to construct new datasets that resemble the actual scenario with our UAV platform.
The software packages needed to reconstruct the datasets for this experiment can be found at \url{https://github.com/britsknguyen/airsim_pathplanner}.

The Building\_99 environment\footnote{\url{https://github.com/microsoft/AirSim/releases/tag/v1.3.1-linux}} is used for this experiment, which features a large indoor foyer area. For this experiment, the following sensors are integrated with the UAV model: one 10 Hz stereo-camera setup, two 10 Hz Lidar sensors, one 400 Hz IMU, two UWB sensors, each with two antennae, and four anchors. The configuration of the UWB network is similar to Section \ref{sec: pub dataset preperation}, except that we set the simulated anchors at the coordinates (0, 0, 11), (-10, 0, 11), (-10, -10, 11), (0, -10, 11). This imitates an installation of these anchors on the ceiling of the building in the real-world scenario. The coordinates of the four UWB ranging nodes in the body frame are set at (0.35, -0.35, 0), (0.35, 0.35, 0), (-0.35, 0.35, 0), (-0.35, -0.35, 0), catered to a bigger platform compared to that in the EuRoC experiment. Finally, for the two Lidars, they are rotated so that one so-called horizontal lidar can effectively scan the surrounding walls, and the other so-called vertical Lidar can scan the floor and the ceiling. The open-sourced A-LOAM software package\footnote{\url{https://github.com/HKUST-Aerial-Robotics/A-LOAM}} is used to provide LOAM estimate from Lidar data. The data from VINS-Fusion and LOAM methods are used as OSL to be fused with UWB and IMU (ORB-SLAM3 are only used for reference as the software package currently does not support publication its data in real-time to ROS).]

Fig. \ref{fig: airsim xyz} presents the trajectories of the UAV in three experiments, along with the estimates from VINS, ORB-SLAM3, LOAM and our proposed method. These trajectories are generated from some fixed setpoints with intended purposes. On the first experiment, we generate a trajectory that goes through all the points in a lattice of cubic cells, and the heading setpoint is also changed by the direction from one setpoint to the next, thus this trajectory is supposed to cover all of the 4 degrees of freedom of the UAV. In the second experiment, the trajectory is similarly set by some setpoints limited to a xy plane. Finally, we generate a trajectory from setpoints on the xz plane, and set the head fixed to the -y direction. This trajectory would most resemble our actual operation in the field test, where a vertical structure will be inspected. From Fig. \ref{fig: airsim xyz}, it can be easily seen that all OSL methods exhibit significant drift over time, while our method does not suffer from this thanks to the installation of some fixed anchors in the environment.


Tab. \ref{tab: airsim ATE} summarizes the RMSE of the estimation errors of our method. Again, we see the advantage of VIRAL method in terms of positional estimation. In addition, the estimate of orientation and velocity estimate also demonstrate the best or second best accuracy compared to other methods. Fig. \ref{fig: airsim vel} shows the velocity estimate from VINS, VIRAL and the grountruth. It can be seen that the velocity estimate follows the groundtruth quite closely, which demonstrates the advantage of our method over previous works that only considered positional states \cite{fang2018model, fang2019graph, wang2017ultra, nguyen2019loosely, nguyen2019tightly}.

\subsection{Field-collected datasets}

\begin{figure}[h]
    \centering
    \includegraphics[width=\linewidth]{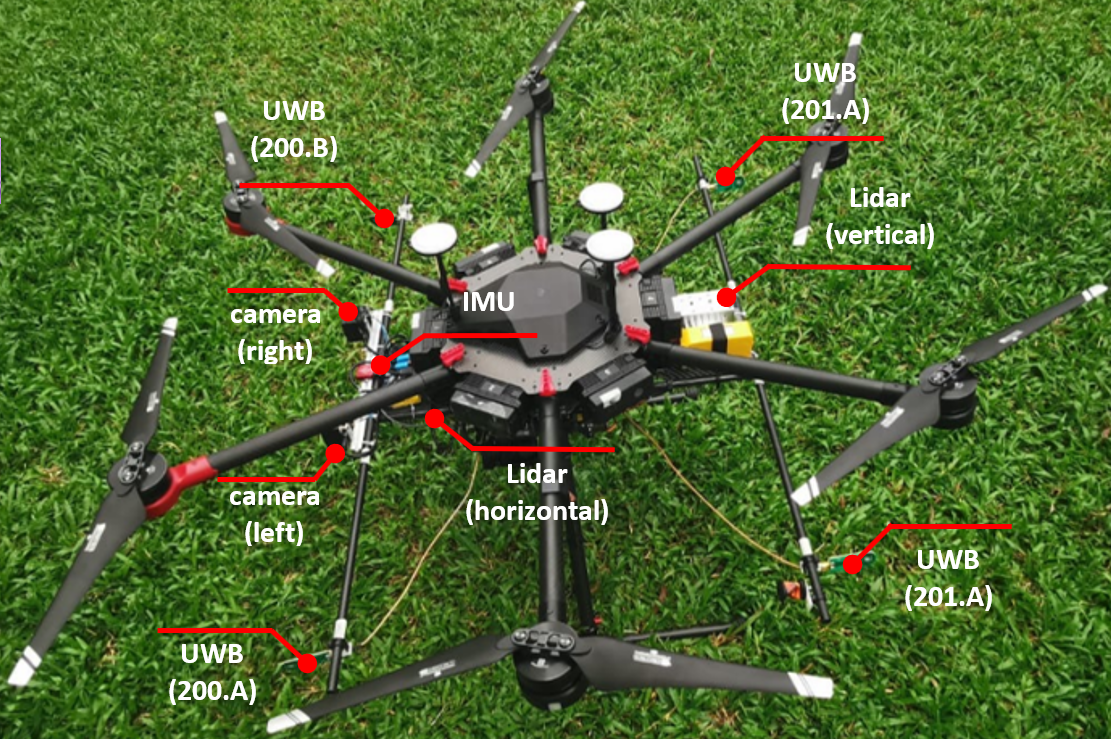}
	\caption{The hardware setup used in our flight tests.}
	\label{fig: hardware setup}
\end{figure}

To further verify the efficacy of the localization scheme, we develop a platform consisting of a camera system\footnote{\url{http://wiki.ros.org/ueye_cam}}, one IMU\footnote{\url{https://www.vectornav.com/products/vn-100}}, two Lidar sensors\footnote{\url{https://ouster.com/products/os1-Lidar-sensor/}}, two UWB modules and three anchors\footnote{\url{https://go.humatics.com/p440}}. The configuration is similar to the AirSim test.
Again, the open-sourced VINS-Fusion and A-LOAM packages are used to generate the VIO and LOAM data for the fusion scheme. Unfortunately, we find that ORB-SLAM3 cannot work with these datasets as the front end process fails to detect reliable features.
A Leica MS60 station\footnote{\url{https://leica-geosystems.com/en-sg/products/total-stations/multistation/leica-nova-ms60}} with millimeter-level accuracy is used to provide groundtruth for the experiment.
All software are run on a small form factor computer\footnote{\url{https://simplynuc.com}} with Intel Core i7-8650U processor. Fig. \ref{fig: hardware setup} presents an overview of our setup.

\begin{figure}[h]
    \centering
    \includegraphics[width=\linewidth]{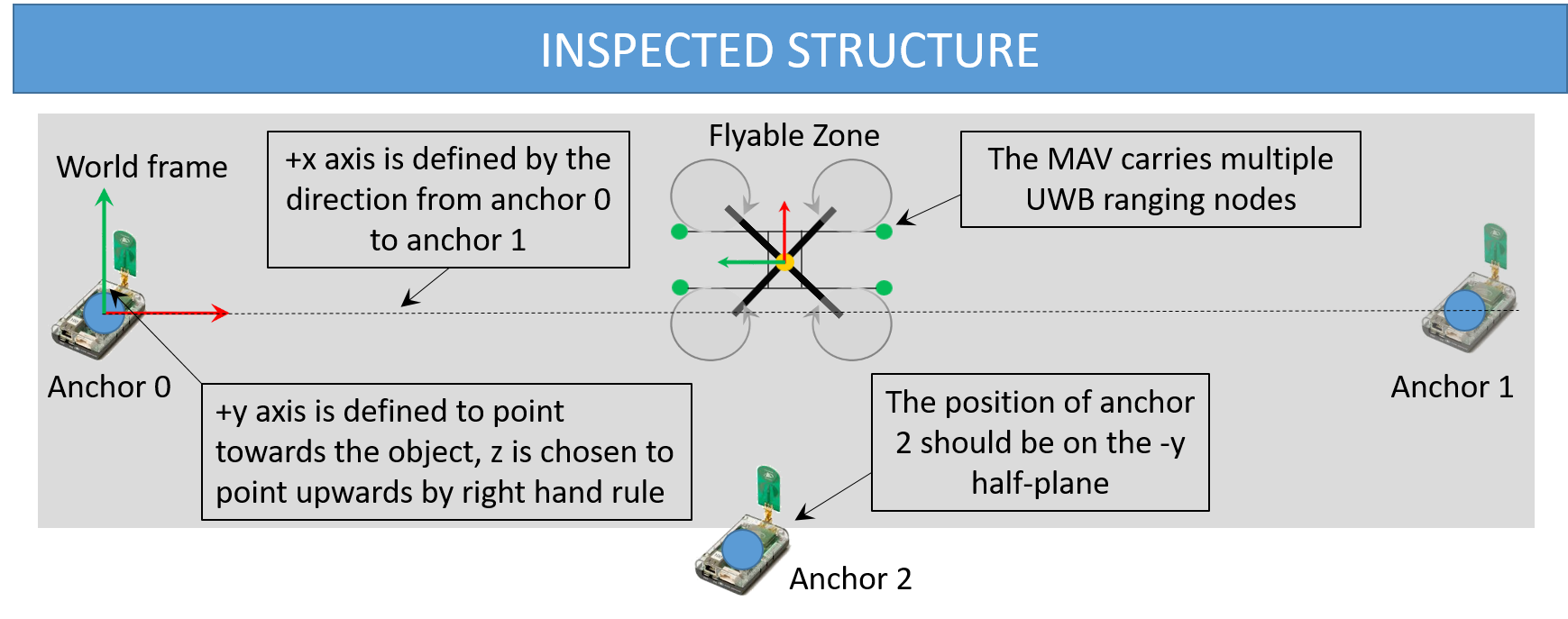}
	\caption{The simple deployment scheme for three anchors.}
	\label{fig: anchor deployment}
\end{figure}

\begin{table}[h]
    {
    \setlength{\tabcolsep}{5pt}
    \renewcommand{\arraystretch}{1.1}
    \centering
	\caption{Estimated coordinates of the three anchors in the four flight tests.}
	\label{tab: anchor coordinates}
    \begin{tabu} to \textwidth {c||c|c|c|c|c|c|c|c|c}
    \hline\hline
    \mr{2}{*}{Test}&
    \mc{3}{c|}{Anc. 0 coord. [m]}  &\mc{3}{c|}{ Anc. 1 coord. [m]} &\mc{3}{c}{Anc. 2 coord. [m]}\\ \cline{2-10} 
               &X &Y &Z   &X &Y &Z  &X &Y &Z\\ \hline
    test\_01   &{0.0}     &{0.0}    &{1.0}
               &{61.55}   &{0.0}    &{1.0}
               &{24.14}   &{-14.15} &{1.0}\\ \hline
    test\_02   &{0.0}     &{0.0}    &{1.0}
               &{39.53}   &{0.0}    &{1.0}
               &{14.28}   &{-9.92}  &{1.0}\\ \hline
    test\_03   &{0.0}     &{0.0}     &{1.0}
               &{44.98}   &{0.0}     &{1.0}
               &{28.46}   &{-9.97}   &{1.0}\\ \hline
    test\_04   &{0.0}     &{0.0}     &{1.0}
               &{12.16}   &{0.0}     &{1.0}
               &{6.08}    &{-11.64}  &{1.0}\\ \hline\hline
    \end{tabu}
    }
\end{table}

\begin{figure*}[h]
    \centering
    \includegraphics[width=\linewidth]{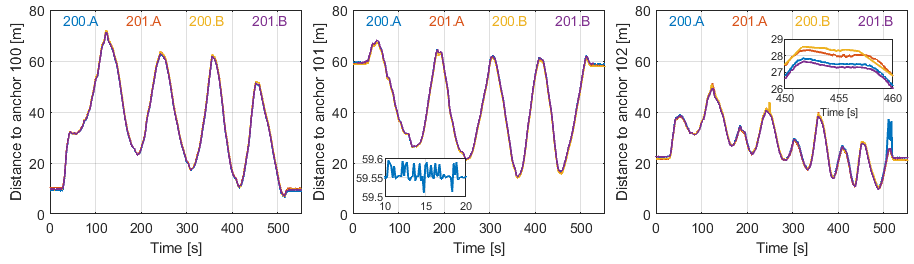}
    \caption{Distance measurements to the anchors from field-collected Dataset 01. From the zoomed-in part of the 200.A$\to$101 plot (middle), the variation of the distance is about 5cm. In the zoomed-in part on the leftmost plot, we can see that the distances between UAV nodes to the anchors are still distinguishable thanks to the significant separation of the UAV nodes on the body frame. We also notice some outliers in the 200.A$\to$102 and 200.B$\to$102 ranging measurements caused by the pilot blocking the anchor during landing. However they are successfully rejected by the algorithm.}
    \label{fig: dist}
\end{figure*}

As having introduced earlier, to keep the problem simple and cost effective, in actual operation we only use three anchors that can be quickly deployed in the field. Fig. \ref{fig: anchor deployment} illustrates our deployment scheme. Specifically, given three anchors 0, 1, 2, we can define the origin of the coordinate system to be at anchor 0. Then the +x axis of the coordinates system can be defined by the direction from anchor 0 to anchor 1. And by choosing for the z axis to point upwards from the ground, we can easily determine the direction of the y axis using the right hand rule, which is intended to point towards the structure that we seek to inspect. Finally, we can place anchor 2 on the -y half plane, further away from the structure than anchor 0 and anchor 1. All of the anchors are supposed to be on a plane parallel to the ground, and their z coordinates are set to be 1m. By using some communication protocol, the UAV can collect the distances between these three anchors, and estimate the remaining unknown coordinates of the anchors, i.e. anchor 1's x coordinate and anchor 2's x and y coordinates. Thanks to this automated self-localization scheme for the anchors, we can quickly set up the anchors at over 60m within 10 to 15 minutes, which mostly comprises of walking time. Otherwise it would be quite a tedious task to measure the coordinates of these anchors, especially at such an extensive distance. The estimated coordinates of the anchors in the flight tests are given in Tab. \ref{tab: anchor coordinates}. Fig. \ref{fig: dist} shows the UWB measurements in one flight test to demonstrate the advantage of the body-offset ranging scheme.

\begin{figure*}[h]
	\centering
	\begin{subfigure}[h]{0.325\linewidth}
        \centering
        \includegraphics[width=\linewidth]{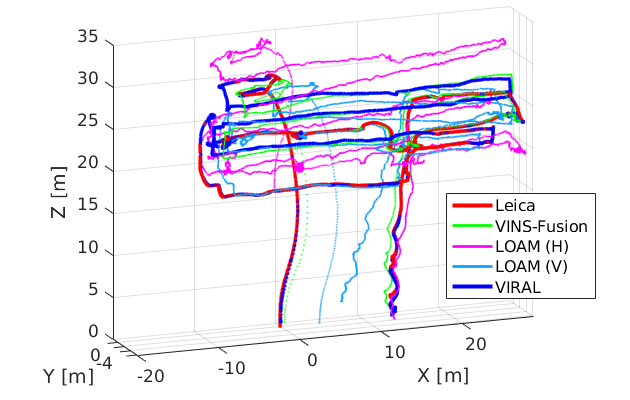}
		\caption{Dataset 01.}
		\label{fig: bca xyz 004}
	\end{subfigure}
    \begin{subfigure}[h]{0.325\linewidth}
        \centering
		\includegraphics[width=\linewidth]{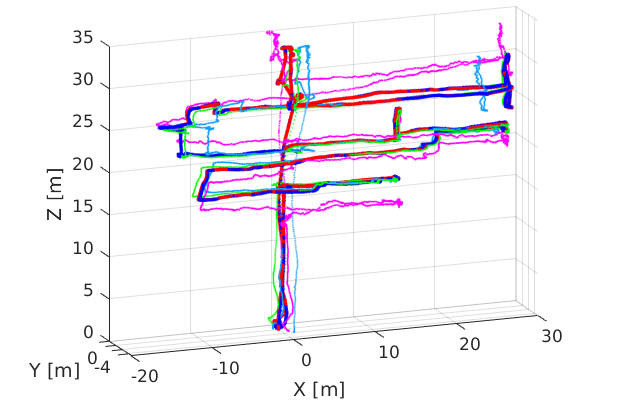}
		\caption{Dataset 02.}
		\label{fig: bca xyz 005}
	\end{subfigure}
	\begin{subfigure}[h]{0.325\linewidth}
        \centering
        \includegraphics[width=\linewidth]{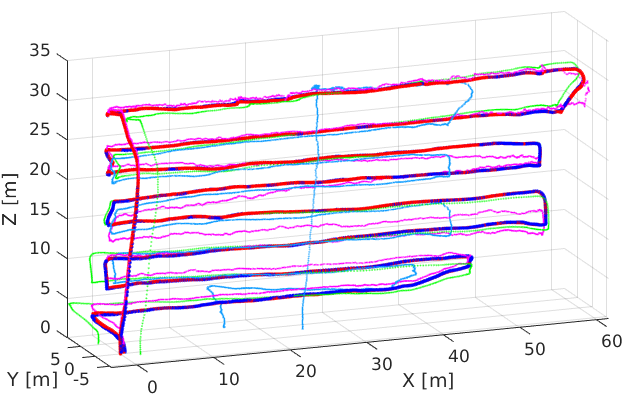}
        \caption{Dataset 03.}
		\label{fig: bca xyz 007}
	\end{subfigure}

	\caption{Trajectories of the UAV and the estimates from multiple methods in the self-collected datasets. In datasets 01 and 02, groundtruth was temporarily lost for some periods. The data in these periods are not used in the error analysis.}  \label{fig: bca xyz}
\end{figure*}

A total of three flight tests were conducted. Video recording of a flight test can be viewed online\footnote{\url{https://youtu.be/2um-n_Wc9-k}}. As having introduced earlier, since our work is meant to deliver a solution for UAV-based inspection application, the UAV's trajectory is different from those in common datasets that are often extensive in x and y directions, but have limited movement in the z direction. In our case, for inspection purposes, the UAV only takes off and moves in a vertical plane parallel to the object. Hence the movements along the x and z directions are extensive, while movement along the y direction is minimal. Fig. \ref{fig: bca xyz} shows the trajectories of the UAV in these flight tests, along with the estimated results.

\begin{figure*}[!h]
	\centering
	 \begin{subfigure}[h]{0.475\linewidth}
        \centering
        \vspace{0.4cm}
        \begin{overpic}[width=\linewidth,
                            ]{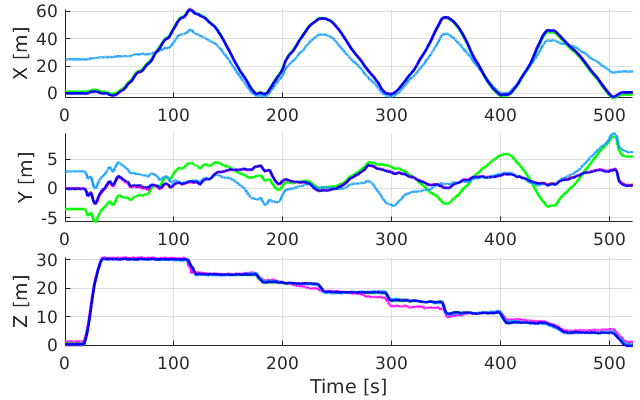}
		\end{overpic}
		\caption{Position estimate from the VIRAL system over time, compared with estimates from OSL systems.}
		\label{fig: facade xyzt}
	\end{subfigure}
	\hfill
    \begin{subfigure}[h]{0.475\linewidth}
		\includegraphics[width=\linewidth]{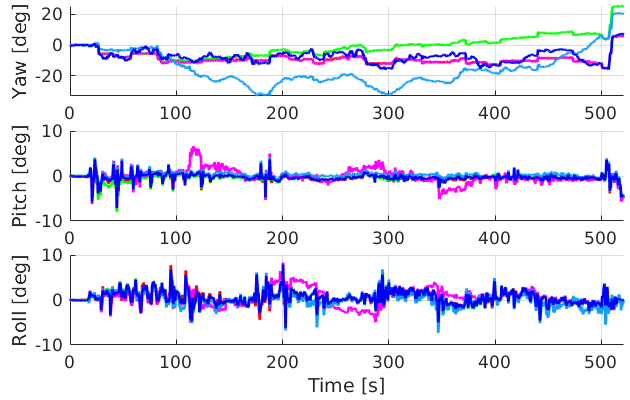}
		\caption{Orientation estimate from the VIRAL system over time, compared with estimates OSL systems.}
		\label{fig: facade yprt}
	\end{subfigure}
	\caption{Position and Orientation estimate from OSL and VIRAL methods in Dataset 03: it can be seen that all OSL methods exhibit significant drift in this scenario, while the VIRAL method provides drift-free estimates despite using only three anchor nodes deployed at the test area.}  \label{fig: facade pos and ori est}
\end{figure*}

\begin{figure*}[h]
	\centering
    \begin{subfigure}[h]{0.475\linewidth}
        \centering
        \begin{overpic}[width=\linewidth,
                            ]{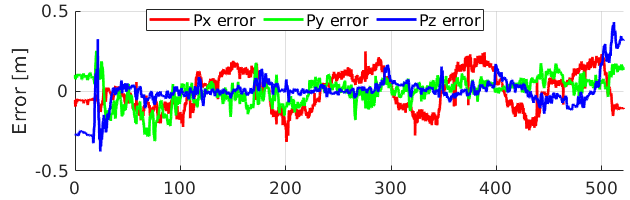}
		\end{overpic}
		\caption{Position estimate error in each direction.}
		\label{fig: facade xyz error}
	\end{subfigure}
    \hfill
	\begin{subfigure}[h]{0.475\linewidth}
		\includegraphics[width=\linewidth]{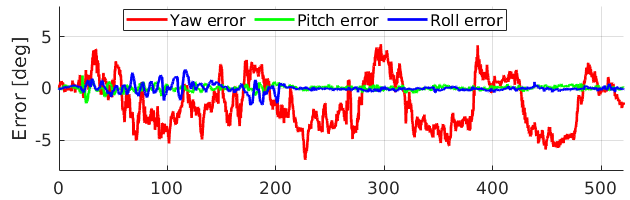}
		\caption{Orientation estimate error in yaw pitch-roll-representation.}
		\label{fig: facade yprt error}
	\end{subfigure}
	\caption{Position and orientation estimation error of VIRAL method in dataset \#03.}  \label{fig: facade error}
\end{figure*}

\begin{table}[h]
    \setlength{\tabcolsep}{3.75pt}
    \centering
    \renewcommand{\arraystretch}{1.1}
	\caption{Translational RMSE VINS-Fusion, LOAM and VIRAL estimates on our collected datasets.}
	\label{tab: collected ATE pos}
    \begin{tabu} to \textwidth {c||c|c|c|c}
    \hline\hline
                        &\mc{4}{c}{$\text{RMSE}_\text{pos}$ [m]}\\\cline{2-5}
    \mr{-2}{*}{Dataset} &VINS   &LOAM (H)  &LOAM (V)  &VIRAL     \\\hline
    01                  &0.8416 &5.4666   &2.9470   &\bf{0.3965} \\
    02                  &0.9381 &2.9086   &1.5804   &\bf{0.2138} \\
    03                  &2.4057 &1.0344   &12.1921  &\bf{0.1692} \\\hline\hline
    \end{tabu}
\end{table}

\begin{table}[h]
    \setlength{\tabcolsep}{3.75pt}
    \centering
    \renewcommand{\arraystretch}{1.1}
	\caption{Rotational RMSE of VINS-Fusion, LOAM and VIRAL estimates on our collected datasets.}
	\label{tab: collected ATE rot}
    \begin{tabu} to \textwidth {c||c|c|c|c}
    \hline\hline
                        &\mc{4}{c}{$\text{RMSE}_\text{rot}$ [deg]}\\\cline{2-5}
    \mr{-2}{*}{Dataset} &VINS       &LOAM (H)    &LOAM (V)  &VIRAL\\\hline
    01                  &{10.6475}  &5.7859      &4.8602    &\bf{3.4666}\\
    02                  &{10.6973}  &7.4157      &5.4855    &\bf{2.6131}\\
    03                  &10.1036    &\bf{1.9316} &11.5245   &2.7682\\\hline\hline
    \end{tabu}
\end{table}

Similar to the experiments previous section, for each dataset, we run an experiment where OSL is fused with UWB and IMU data. Since the transform between ground truth coordinate frame and the local frame of reference is not known, we transform the esmiates using the method in \cite{zhang2018tutorial}, which finds the yaw and translation that minimize the position estimation error of each trajectory, and report these errors Tab. \ref{tab: collected ATE pos}. Moreover, since the leica ground station does not provide orientation measurements, we use the data from the onboard attitude and heading reference system (AHRS) of the drone as ground truth, which is trusted to be accurate as it fuses magnetometer measurements and no interference was detected at the test location on this day. The orientation estimation errors are reported in Tab. \ref{tab: collected ATE rot}. Fig. \ref{fig: facade pos and ori est}, Fig. \ref{fig: facade error} show the detailed plots of the position and orientation estimates of one of the experiments.

From the results in Tab. \ref{tab: collected ATE pos} and Tab. \ref{tab: collected ATE rot}, we can reaffirm benefit of the UWB-IMU-OSL fusion scheme, whose positional error is much smaller than the OSL methods. In Fig. \ref{fig: facade xyz error}, we can see that the position error is bounded. The orientation estimate also performs quite well, and Fig. \ref{fig: facade yprt error} also shows that the orientation is bounded. Compared to the other experiments on the EuRoC and AirSim datasets, the positional error is slightly larger. This can be attributed to some small error in the anchor self-localization procedure, as the anchors may not be exactly on the same level. Also, some biases in the distance measurements due to extension cable may also affect the accuracy of the UWB measurements. Nevertheless, as can be seen on Fig. \ref{fig: facade pos and ori est} that while all of the OSL estimates exhibit significant drift over the operation range, our sensor fusion's estimation error remains bounded, which is the most important feature required for this work.

\section{Conclusion} \label{sec: conclusion}

In this paper we have developed an optimization based sensor fusion framework for visual, inertial, ranging and Lidar measurements. Detailed mathematical models for the construction of the cost function and the on-manifold optimization procedure have been carefully derived. The framework has been successfully implemented with open-source ceres solver and ROS framework. We have also conducted experiments to verify the efficacy and effectiveness of the sensor fusion scheme on multiple datasets, and shown that via the use of some anchors that can be quickly deployed on the field, the issues of OSL drift and misaligned frame of references can be effectively overcome. Moreover, results on estimation of position, orientation, and velocity have also been presented to show the comprehensiveness of the estimator. Via the simulation and experimental results, it can be seen that the VIRAL scheme is reliable, flexible, accurate and has a great potential in many real-world applications.

\balance
\bibliographystyle{IEEEtran}
\bibliography{references}

\begin{thebibliography}{10}
\providecommand{\url}[1]{#1}
\csname url@samestyle\endcsname
\providecommand{\newblock}{\relax}
\providecommand{\bibinfo}[2]{#2}
\providecommand{\BIBentrySTDinterwordspacing}{\spaceskip=0pt\relax}
\providecommand{\BIBentryALTinterwordstretchfactor}{4}
\providecommand{\BIBentryALTinterwordspacing}{\spaceskip=\fontdimen2\font plus
\BIBentryALTinterwordstretchfactor\fontdimen3\font minus
  \fontdimen4\font\relax}
\providecommand{\BIBforeignlanguage}[2]{{%
\expandafter\ifx\csname l@#1\endcsname\relax
\typeout{** WARNING: IEEEtran.bst: No hyphenation pattern has been}%
\typeout{** loaded for the language `#1'. Using the pattern for}%
\typeout{** the default language instead.}%
\else
\language=\csname l@#1\endcsname
\fi
#2}}
\providecommand{\BIBdecl}{\relax}
\BIBdecl

\bibitem{shen20113d}
S.~Shen, N.~Michael, and V.~Kumar, ``3d indoor exploration with a
  computationally constrained mav,'' in \emph{Robotics: Science and Systems},
  2011.

\bibitem{bloesch2015robust}
M.~Bloesch, S.~Omari, M.~Hutter, and R.~Siegwart, ``Robust visual inertial
  odometry using a direct ekf-based approach,'' in \emph{Intelligent Robots and
  Systems (IROS), 2015 IEEE/RSJ International Conference on}.\hskip 1em plus
  0.5em minus 0.4em\relax IEEE, 2015, pp. 298--304.

\bibitem{engel2014lsd}
J.~Engel, T.~Sch{\"o}ps, and D.~Cremers, ``Lsd-slam: Large-scale direct
  monocular slam,'' in \emph{European conference on computer vision}.\hskip 1em
  plus 0.5em minus 0.4em\relax Springer, 2014, pp. 834--849.

\bibitem{weinstein2018visual}
A.~Weinstein, A.~Cho, G.~Loianno, and V.~Kumar, ``Visual inertial odometry
  swarm: An autonomous swarm of vision-based quadrotors,'' \emph{IEEE Robotics
  and Automation Letters}, vol.~3, no.~3, pp. 1801--1807, 2018.

\bibitem{forster2014svo}
C.~Forster, M.~Pizzoli, and D.~Scaramuzza, ``Svo: Fast semi-direct monocular
  visual odometry,'' in \emph{2014 IEEE International Conference on Robotics
  and Automation (ICRA)}.\hskip 1em plus 0.5em minus 0.4em\relax IEEE, 2014,
  pp. 15--22.

\bibitem{qin2017vins}
T.~Qin, P.~Li, and S.~Shen, ``Vins-mono: A robust and versatile monocular
  visual-inertial state estimator,'' \emph{IEEE Transactions on Robotics},
  vol.~34, no.~4, pp. 1004--1020, 2018.

\bibitem{mur2017orb}
R.~Mur-Artal and J.~D. Tard{\'o}s, ``Orb-slam2: An open-source slam system for
  monocular, stereo, and rgb-d cameras,'' \emph{IEEE Transactions on Robotics},
  vol.~33, no.~5, pp. 1255--1262, 2017.

\bibitem{zhang2018laser}
J.~Zhang and S.~Singh, ``Laser--visual--inertial odometry and mapping with high
  robustness and low drift,'' \emph{Journal of Field Robotics}, vol.~35, no.~8,
  pp. 1242--1264, 2018.

\bibitem{ye2019tightly}
H.~Ye, Y.~Chen, and M.~Liu, ``Tightly coupled 3d lidar inertial odometry and
  mapping,'' in \emph{2019 IEEE International Conference on Robotics and
  Automation (ICRA)}.\hskip 1em plus 0.5em minus 0.4em\relax IEEE, 2019.

\bibitem{shan2020liosam}
T.~Shan, B.~Englot, D.~Meyers, W.~Wang, C.~Ratti, and R.~Daniela, ``Lio-sam:
  Tightly-coupled lidar inertial odometry via smoothing and mapping,'' in
  \emph{IEEE/RSJ International Conference on Intelligent Robots and Systems
  (IROS)}.\hskip 1em plus 0.5em minus 0.4em\relax IEEE, 2020.

\bibitem{suleiman2019navion}
A.~Suleiman, Z.~Zhang, L.~Carlone, S.~Karaman, and V.~Sze, ``Navion: A 2-mw
  fully integrated real-time visual-inertial odometry accelerator for
  autonomous navigation of nano drones,'' \emph{IEEE Journal of Solid-State
  Circuits}, vol.~54, no.~4, pp. 1106--1119, 2019.

\bibitem{qin2018online}
T.~Qin and S.~Shen, ``Online temporal calibration for monocular visual-inertial
  systems,'' in \emph{2018 IEEE/RSJ International Conference on Intelligent
  Robots and Systems (IROS)}.\hskip 1em plus 0.5em minus 0.4em\relax IEEE,
  2018, pp. 3662--3669.

\bibitem{guo2016ultra}
K.~Guo, Z.~Qiu, C.~Miao, A.~H. Zaini, C.-L. Chen, W.~Meng, and L.~Xie,
  ``Ultra-wideband-based localization for quadcopter navigation,''
  \emph{Unmanned Systems}, vol.~4, no.~01, pp. 23--34, 2016.

\bibitem{nguyen2016ultra}
T.-M. Nguyen, A.~H. Zaini, K.~Guo, and L.~Xie, ``An ultra-wideband-based
  multi-uav localization system in gps-denied environments,'' in
  \emph{International Micro Air Vehicle Competition and Conference 2016},
  Beijing, China, Oct 2016, pp. 56--61.

\bibitem{tiemann2017scalable}
J.~Tiemann and C.~Wietfeld, ``Scalable and precise multi-uav indoor navigation
  using tdoa-based uwb localization,'' in \emph{2017 International Conference
  on Indoor Positioning and Indoor Navigation (IPIN)}.\hskip 1em plus 0.5em
  minus 0.4em\relax IEEE, 2017, pp. 1--7.

\bibitem{mueller2015fusing}
M.~W. Mueller, M.~Hamer, and R.~D'Andrea, ``Fusing ultra-wideband range
  measurements with accelerometers and rate gyroscopes for quadrocopter state
  estimation,'' in \emph{2015 IEEE International Conference on Robotics and
  Automation (ICRA)}.\hskip 1em plus 0.5em minus 0.4em\relax IEEE, 2015, pp.
  1730--1736.

\bibitem{li2018accurate}
J.~Li, Y.~Bi, K.~Li, K.~Wang, F.~Lin, and B.~M. Chen, ``Accurate 3d
  localization for mav swarms by uwb and imu fusion,'' in \emph{2018 14th IEEE
  International Conference on Control and Automation (ICCA)}.\hskip 1em plus
  0.5em minus 0.4em\relax IEEE, 2018, pp. 100--105.

\bibitem{fang2018model}
X.~Fang, C.~Wang, T.-M. Nguyen, and L.~Xie, ``Model-free approach for sensor
  network localization with noisy distance measurement,'' in \emph{2018 15th
  International Conference on Control, Automation, Robotics and Vision
  (ICARCV)}.\hskip 1em plus 0.5em minus 0.4em\relax IEEE, 2018, pp. 1973--1978.

\bibitem{fang2019graph}
X.~{Fang}, C.~{Wang}, T.-M. {Nguyen}, and L.~{Xie}, ``Graph optimization
  approach to range-based localization,'' \emph{IEEE Transactions on Systems,
  Man, and Cybernetics: Systems}, pp. 1--12, 2020.

\bibitem{wang2017ultra}
C.~Wang, H.~Zhang, T.-M. Nguyen, and L.~Xie, ``{Ultra-Wideband Aided Fast
  Localization and Mapping System},'' in \emph{2017 IEEE/RSJ International
  Conference on Intelligent Robots and Systems (IROS)}.\hskip 1em plus 0.5em
  minus 0.4em\relax IEEE, 2017.

\bibitem{nguyen2019loosely}
\BIBentryALTinterwordspacing
T.~H. Nguyen, T.-M. Nguyen, M.~Cao, and L.~Xie, ``{Loosely-Coupled
  Ultra-Wideband-Aided Scale Correction for Monocular Visual Odometry},''
  \emph{Unmanned Systems}, vol.~08, no.~02, pp. 179--190, 2020. [Online].
  Available:
  \url{https://www.worldscientific.com/doi/10.1142/S2301385020500119}
\BIBentrySTDinterwordspacing

\bibitem{nguyen2019tightly}
T.~H. Nguyen, T.-M. Nguyen, and L.~Xie, ``Tightly-coupled single-anchor
  ultra-wideband-aided monocular visual odometry system,'' in \emph{2020 IEEE
  International Conference on Robotics and Automation (ICRA).}\hskip 1em plus
  0.5em minus 0.4em\relax Paris, France: IEEE, 2020.

\bibitem{nguyen2020tightly}
------, ``{Tightly-Coupled Ultra-wideband-Aided Monocular Visual SLAM with
  Degenerate Anchor Configurations},'' \emph{Autonomous Robots (Under Review)},
  2020.

\bibitem{cao2020vir}
Y.~Cao and G.~Beltrame, ``Vir-slam: Visual, inertial, and ranging slam for
  single and multi-robot systems,'' \emph{arXiv preprint arXiv:2006.00420},
  2020.

\bibitem{song2019uwb}
Y.~Song, M.~Guan, W.~P. Tay, C.~L. Law, and C.~Wen, ``Uwb/lidar fusion for
  cooperative range-only slam,'' in \emph{2019 International Conference on
  Robotics and Automation (ICRA)}.\hskip 1em plus 0.5em minus 0.4em\relax IEEE,
  2019, pp. 6568--6574.

\bibitem{paredes20183d}
J.~Paredes, F.~{\'A}lvarez, T.~Aguilera, and J.~Villadangos, ``3d indoor
  positioning of uavs with spread spectrum ultrasound and time-of-flight
  cameras,'' \emph{Sensors}, vol.~18, no.~1, p.~89, 2018.

\bibitem{mautz2012indoor}
\BIBentryALTinterwordspacing
R.~Mautz, ``Indoor positioning technologies,'' \emph{ETH Zurich's Research
  Collection}, 2012. [Online]. Available:
  \url{https://doi.org/10.3929/ethz-a-007313554}
\BIBentrySTDinterwordspacing

\bibitem{kummerle2011g}
R.~K{\"u}mmerle, G.~Grisetti, H.~Strasdat, K.~Konolige, and W.~Burgard, ``g 2
  o: A general framework for graph optimization,'' in \emph{2011 IEEE
  International Conference on Robotics and Automation (ICRA)}.\hskip 1em plus
  0.5em minus 0.4em\relax IEEE, 2011, pp. 3607--3613.

\bibitem{ceres-solver}
S.~Agarwal, K.~Mierle, and Others, ``Ceres solver,''
  \url{http://ceres-solver.org}.

\bibitem{qin2018vins}
T.~Qin, P.~Li, and S.~Shen, ``Vins-mono: A robust and versatile monocular
  visual-inertial state estimator,'' \emph{IEEE Transactions on Robotics},
  vol.~34, no.~4, pp. 1004--1020, 2018.

\bibitem{campos2020orb}
C.~Campos, R.~Elvira, J.~J.~G. Rodr{\'\i}guez, J.~M. Montiel, and J.~D.
  Tard{\'o}s, ``Orb-slam3: An accurate open-source library for visual,
  visual-inertial and multi-map slam,'' \emph{arXiv preprint arXiv:2007.11898},
  2020.

\bibitem{chirikjian2011stochastic}
G.~S. Chirikjian, \emph{Stochastic Models, Information Theory, and Lie Groups,
  Volume 2: Analytic Methods and Modern Applications}.\hskip 1em plus 0.5em
  minus 0.4em\relax Springer Science \& Business Media, 2011, vol.~2.

\bibitem{sola2018micro}
J.~Sola, J.~Deray, and D.~Atchuthan, ``A micro lie theory for state estimation
  in robotics,'' \emph{arXiv preprint arXiv:1812.01537}, 2018.

\bibitem{huang2020online}
W.~Huang, H.~Liu, and W.~Wan, ``An online initialization and self-calibration
  method for stereo visual-inertial odometry,'' \emph{IEEE Transactions on
  Robotics}, 2020.

\bibitem{forster2016manifold}
C.~Forster, L.~Carlone, F.~Dellaert, and D.~Scaramuzza, ``On-manifold
  preintegration for real-time visual--inertial odometry,'' \emph{IEEE
  Transactions on Robotics}, vol.~33, no.~1, pp. 1--21, 2016.

\bibitem{murorb2mono}
R.~Mur-Artal, J.~M.~M. Montiel, and J.~D. Tard\'os, ``{ORB-SLAM}: a versatile
  and accurate monocular {SLAM} system,'' \emph{IEEE Transactions on Robotics},
  vol.~31, no.~5, pp. 1147--1163, 2015.

\bibitem{eckenhoff2019closed}
K.~Eckenhoff, P.~Geneva, and G.~Huang, ``Closed-form preintegration methods for
  graph-based visual--inertial navigation,'' \emph{The International Journal of
  Robotics Research}, vol.~38, no.~5, pp. 563--586, 2019.

\bibitem{burri2016euroc}
M.~Burri, J.~Nikolic, P.~Gohl, T.~Schneider, J.~Rehder, S.~Omari, M.~W.
  Achtelik, and R.~Siegwart, ``The euroc micro aerial vehicle datasets,''
  \emph{The International Journal of Robotics Research}, vol.~35, no.~10, pp.
  1157--1163, 2016.

\bibitem{shah2018airsim}
S.~Shah, D.~Dey, C.~Lovett, and A.~Kapoor, ``Airsim: High-fidelity visual and
  physical simulation for autonomous vehicles,'' in \emph{Field and service
  robotics}.\hskip 1em plus 0.5em minus 0.4em\relax Springer, 2018, pp.
  621--635.

\bibitem{zhang2018tutorial}
Z.~Zhang and D.~Scaramuzza, ``A tutorial on quantitative trajectory evaluation
  for visual (-inertial) odometry,'' in \emph{2018 IEEE/RSJ International
  Conference on Intelligent Robots and Systems (IROS)}.\hskip 1em plus 0.5em
  minus 0.4em\relax IEEE, 2018, pp. 7244--7251.

\end{thebibliography}

\end{document}